\documentclass[a4paper,fleqn]{cas-dc}

\usepackage[numbers]{natbib}
\usepackage{amsmath,amsfonts,dsfont}
\usepackage{amsthm}
\usepackage[noend]{algorithmic}
\usepackage{algorithm}
\usepackage{cleveref}
\usepackage{tabularx}
\usepackage{array}
\usepackage{multirow}

\DeclareMathOperator*{\argmin}{argmin} 
\DeclareMathOperator*{\argmax}{argmax}
\DeclareFontFamily{U}{mathx}{}
\DeclareFontShape{U}{mathx}{m}{n}{<-> mathx10}{}
\DeclareSymbolFont{mathx}{U}{mathx}{m}{n}

\newtheorem{assumption}{Assumption}

\begin{document}
\let\WriteBookmarks\relax
\def\floatpagepagefraction{1}
\def\textpagefraction{.001}
\shorttitle{Model-Based Reinforcement Learning Under Distribution Shifts}
\shortauthors{D. Garces et~al.}

\title [mode = title]{Model-Based Reinforcement Learning for Heterogeneous Multi-Robot Task Assignment Under Distribution Shifts}

\author[1]{Daniel Garces}[type=editor,
                        orcid=0000-0002-4161-0265]
\cormark[1]
\ead{dgarces@g.harvard.edu}

\credit{Conceptualization, Formal Analysis, Investigation, Methodology, Software, Validation, Visualization, Writing - original draft, Writing - review and editing}

\affiliation[1]{organization={John A. Paulson School Of Engineering And Applied Sciences, Harvard University},
                addressline={150 Western Avenue}, 
                city={Boston},
                postcode={02134}, 
                state={MA},
                country={USA}}

\author[2]{Sara Castro}

\credit{Conceptualization, Data curation, Methodology, Writing - review and editing}

\affiliation[2]{organization={Harvard Medical School},
                addressline={77 Avenue Louis Pasteur}, 
                city={Boston},
                postcode={02215},
                state={MA},
                country={USA}}

\author[2,3]{Adrian Haimovich}

\credit{Data curation, Supervision}

\affiliation[3]{organization={Beth Israel Deaconess Medical Center},
                addressline={330 Brookline Avenue},
                city={Boston},
                postcode={02215},
                state={MA},
                country={USA}}

\author[4]{Byron Crowe}
\affiliation[4]{organization={Stanford University School of Medicine},
                addressline={300 Pasteur Drive},
                city={Stanford},
                postcode={94305},
                state={CA},
                country={USA}}

\credit{Conceptualization, Data curation, Supervision}

\author[1]{Stephanie Gil}

\credit{Funding Acquisition, Supervision, Writing - review and editing}

\cortext[cor1]{Corresponding author}

\begin{abstract}
Heterogeneous multi-robot service systems must assign requests to compatible robots, construct feasible schedules, and adapt as new tasks arrive online. Historical data can help anticipate future demand, but relying too heavily on inaccurate predictions can degrade performance under distribution shifts. We develop a prediction-aware adaptive rollout framework for heterogeneous multi-robot task assignment with scheduled and real-time requests. The problem is formulated as a finite-horizon stochastic dynamic program incorporating robot-task compatibility, ordered service requirements, routing constraints, service windows, and end-of-horizon return requirements. The proposed policy evaluates current assignments using sampled future request scenarios while restricting immediate commitments to requests already observed. To enable online use, the framework combines pruned candidate controls, wait actions, and an interaction-aware base policy for efficient future-cost estimation. Robustness to forecast error is provided by adaptively reweighting predicted requests based on recent prediction mismatch and selectively re-optimizing assigned but unstarted requests. We also introduce a historical-data-driven procedure for selecting the heterogeneous fleet composition before deployment. In a case study using real nursing-task requests from hospital inpatient floors, the proposed approach achieves near-complete service and reduces serviced-request wait times relative to reactive, token-passing, prediction-positioning, and myopic greedy baselines, with the largest improvements in tail-delay metrics.
\end{abstract}

\begin{highlights}
\item Online task assignment for heterogeneous robots with service windows.
\item Prediction-aware rollout uses future requests without committing to them.
\item Adaptive weights reduce the impact of unreliable demand forecasts.
\item Selective re-optimization revises assigned but unstarted requests.
\item Hospital case study shows lower wait times and near-complete service.
\end{highlights}


\maketitle

\section{Introduction}
\label{sec:introduction}

Autonomous robot teams are increasingly being deployed in service environments where spatially distributed requests must be completed under timing, routing, and  compatibility constraints. Examples include monitoring and maintenance \cite{Nishi2005TRO, Zheng2005TRO, Bopardikar2014TRO, Testa2022TRO}, warehouse and logistics operations \cite{Ham01022021, Sorbelli2022TRO}, last-mile delivery and transportation services \cite{Matthew2015, Reed2022, Lee2023, Camisa2023TRO, Jokinen2011, Ongel2019, SHAHEEN2019, Kondor2022}, UAV resupply and freight management \cite{Arribas2023TRO, Xidias2022}, and service coordination during emergency or pandemic responses \cite{Fu2023TRO}. In these settings, robot teams rarely execute a fixed set of tasks known completely in advance. Instead, they must repeatedly adapt assignments and schedules as new requests arrive, robot availability changes, and operating conditions deviate from historical expectations.

This paper studies online task assignment and scheduling for heterogeneous multi-robot teams operating over a finite horizon. The team must serve both scheduled requests, known before execution, and real-time requests, revealed sequentially during operation. Requests may require ordered visits to one or more service locations and must satisfy timing, routing, and compatibility constraints. Robots may differ in their service capabilities, travel times, operational status, and maintenance requirements. The resulting planning problem is to assign observed requests to robots while preserving enough scheduling flexibility to respond effectively to uncertain future demand.

A key challenge is that online assignment decisions have delayed consequences. Assigning a robot to a currently observed request changes the robot's future location, availability, and ability to serve requests that may arrive later. Predictive models can help by using historical data to generate possible future request scenarios, allowing the planner to estimate the opportunity cost of current commitments. However, operating conditions may change during deployment: the number, timing, location, or type of real-time requests may differ from the historical patterns used to generate predictions. When this occurs, a planner that continues to rely heavily on inaccurate forecasts may reserve robots unnecessarily, delay urgent observed requests, or overcommit resources to schedules that are poorly matched to the demand actually being realized. This paper addresses the question of how an online heterogeneous multi-robot planner should use predictive demand information when that information is useful on average but may become unreliable during operation.

Existing approaches only partially address this issue. Reactive online assignment methods \cite{Verma2024,Li2021,Zhao2022,Das2015,Bai2024Cluster,Bai2024Impact,Ferreira2022,Wei2020,Kalempa2021} can respond to newly observed requests, but they are typically myopic and do not explicitly reason about the future consequences of current commitments. Prediction-based methods \cite{Wang2022, Gao2023, Park2022RL,shida2024,zhang2025graph,Dai2025} can anticipate likely future demand, but they often lack mechanisms for adjusting the influence of predictions when recent observations indicate forecast mismatch. Offline deterministic scheduling methods \cite{LYU2019, Zuo2021, WANG2023IC} assume that the relevant requests are known in advance, while exact stochastic dynamic programming formulations \cite{BERBEGLIA20108, LOWALEKAR201871} are computationally intractable for the heterogeneous online setting considered here if no approximations are used. These limitations motivate an approximate online planning method that can use lookahead to anticipate future demand when predictions are reliable, reduce the influence of predictions when recent observations indicate forecast mismatch, and avoid overcommitting robots when future demand remains uncertain.

Rollout-based lookahead \cite{bertsekas2019reinforcement, bertsekas2020rollout} provides a natural way to construct such an approximation. Because current assignments affect future robot availability, sampled future request scenarios can be used to estimate the opportunity cost of assigning a robot to a currently observed request, and candidate decisions can be compared according to their immediate and simulated future costs. However, deploying rollout in this setting introduces significant computational challenges. Each candidate decision is itself a schedule update: the planner must decide not only which robot should serve each known request, but also where that request should be inserted in the robot's route and when each service should occur. As a result, the feasible schedule space grows combinatorially with the number of robots, requests, compatibility relations, and insertion positions, making exhaustive action enumeration impractical. In addition, each candidate schedule must be evaluated by simulating future decisions across sampled request scenarios. These simulations must be efficient enough to run repeatedly online, yet detailed enough to preserve the dominant interactions among robot compatibility, graph-based routing, service timing, and robot availability. Beyond the rollout computation itself, the usefulness of these simulated policies also depends on the initial heterogeneous robot composition: if the team lacks sufficient compatible capacity for bottleneck request types, then even an effective online scheduling policy may be unable to maintain feasible service.

Accurate future-cost estimation is also difficult because the request scenarios used in rollout are only approximations of future operating conditions. Predictions generated from historical data may be informative on average, but the realized number, timing, location, or type of real-time requests can change during deployment. When this occurs, simulated future costs based on outdated or inaccurate forecasts may misrepresent the true opportunity cost of current assignments. Moreover, forecast errors may have already influenced the current schedule before they are detected. For this reason, future-cost estimates should reflect the planner's current confidence in the predicted request scenarios, rather than treating all predictions as equally reliable throughout the operating horizon. The planner must also be able to revise unexecuted assignments when new observations indicate that the current schedule was shaped by unreliable forecasts, while preserving commitments to services that have already begun.

We address these challenges with a prediction-aware adaptive rollout framework for heterogeneous multi-robot task assignment and scheduling. The central idea is to treat predictions as uncertain evidence whose influence should vary over time, rather than as fixed information that should be trusted uniformly throughout the operating horizon. At each decision time, sampled future request scenarios are used to estimate the downstream cost of candidate scheduling decisions. Their contribution to these estimates is modulated by confidence weights derived from recent discrepancies between predicted and observed demand. In this way, the planner can use predictions to make anticipatory decisions when they are informative, while reducing their influence when realized operating conditions suggest that the forecast is unreliable.

The framework combines this adaptive use of predictions with computational mechanisms needed for online deployment. A pruned action-generation procedure constructs a tractable set of candidate schedule modifications instead of enumerating the full feasible schedule space. Explicit wait actions allow the planner to preserve robot capacity when committing immediately to an observed request may be undesirable under uncertain future demand. An interaction-aware base policy is used inside rollout simulations to approximate future assignment and routing decisions efficiently while preserving the main coupling effects among timing, routing, compatibility, and robot availability. A selective re-optimization mechanism can return eligible assigned but unstarted requests to the pending set when new observations indicate that the current schedule has become undesirable, while already-started services remain fixed. As a supporting pre-deployment component, we also develop a historical-data-driven fleet-sizing procedure that selects an initial heterogeneous robot composition for which the assignment routines achieve a target empirical feasibility level on representative operating days.

\begin{figure*}
\centering
\includegraphics[width=\linewidth]{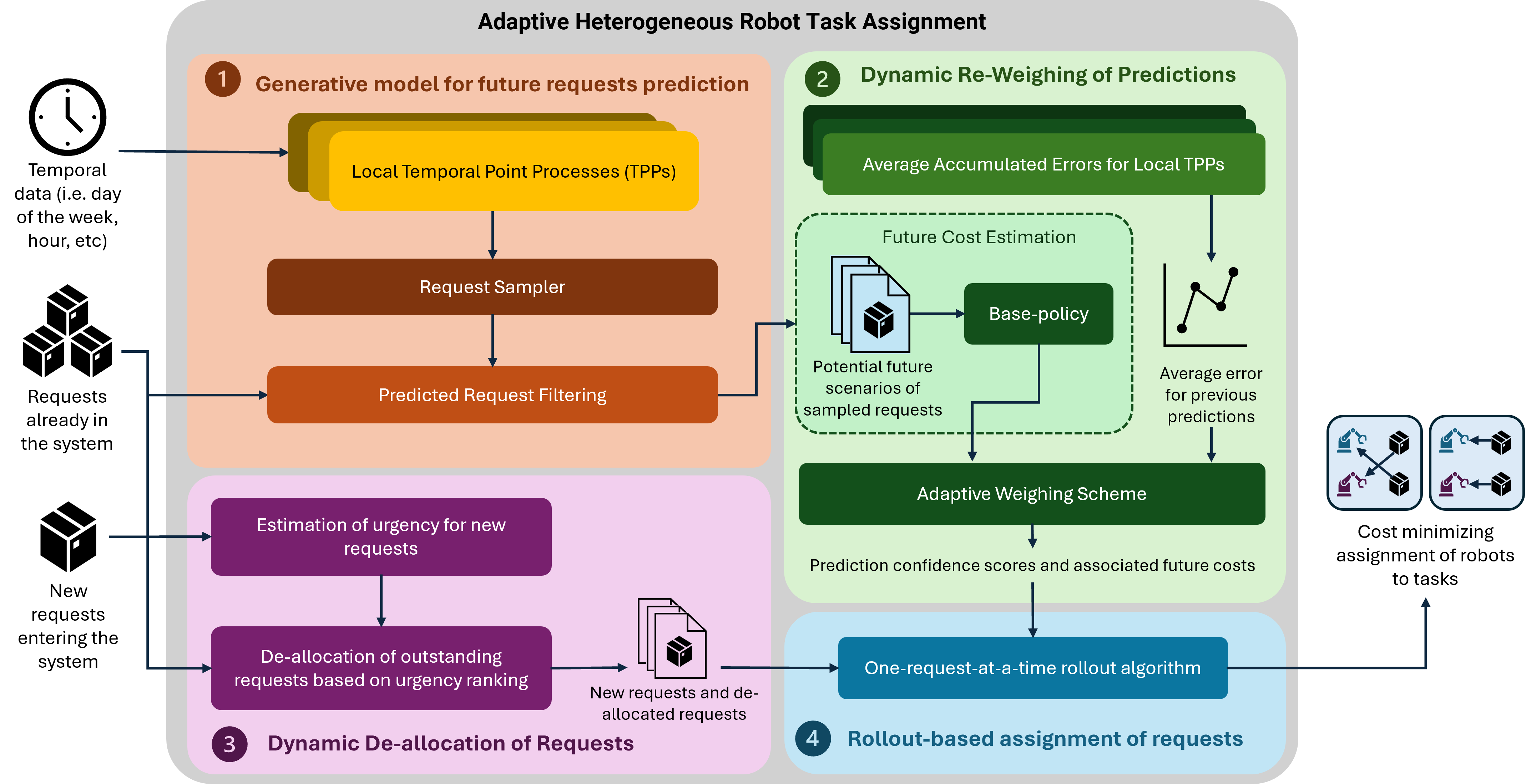}
\caption{\small{Overview of the proposed prediction-aware adaptive rollout framework for heterogeneous multi-robot task assignment. Local temporal point process models generate sampled future request scenarios, which are filtered to remove requests that are already observed or scheduled. Recent forecast errors are used to dynamically reweight predicted requests, producing context-dependent prediction-confidence scores and weighted future-cost estimates. Newly observed urgent requests can trigger re-optimization by returning eligible assigned but unstarted requests to the pending set. The rollout-based assignment module evaluates candidate assignments for observed and re-optimized requests using the weighted future-cost estimates and selects schedules for the heterogeneous robot team.}}
\label{fig:general_overview}
\end{figure*}

An overview of the proposed adaptive rollout method is shown in \Cref{fig:general_overview}. The framework integrates prediction, adaptive forecast reweighting, selective re-optimization, and rollout-based schedule evaluation. Together, these components produce an online planner that is anticipatory when predictions are useful, adaptive when operating conditions change, computationally tractable for repeated online decisions, and less prone to overcommitting robot capacity when future demand is uncertain.

The main contributions of this paper are as follows:
\begin{enumerate}
\item We formulate online heterogeneous multi-robot task assignment and scheduling with scheduled and real-time requests as a finite-horizon stochastic dynamic program that captures robot-request compatibility, ordered service requirements, timing constraints, graph-based routing, and end-of-horizon feasibility.
\item We develop a prediction-aware adaptive rollout framework for online scheduling under uncertain demand. The framework uses sampled future request scenarios to estimate the opportunity cost of current decisions, dynamically adjusts their influence based on recent forecast accuracy, and selectively re-optimizes assigned but unstarted requests when operating conditions deviate from predictions. Pruned action generation, explicit wait actions, and an interaction-aware base policy make the resulting rollout computation practical for repeated online use.
\item We develop a historical-data-driven fleet-sizing procedure for selecting an initial heterogeneous robot composition that achieves a target empirical feasibility level on representative operating days while accounting for robot-request compatibility and bottleneck service requirements.
\item We evaluate the framework using real nursing-task requests from hospital inpatient floors at Beth Israel Deaconess Medical Center. Compared with reactive dispatching, token-passing, prediction-positioning, and myopic greedy baselines, the proposed method achieves near-complete service while reducing both average and upper-tail request waiting times. Ablation studies further demonstrate the complementary roles of adaptive prediction reweighting and selective re-optimization.
\end{enumerate}

The remainder of the paper is organized as follows. \Cref{sec:related_work} reviews related work on heterogeneous online task allocation, fleet sizing, request-sequence prediction, rollout-based replanning, and robustness under distribution shift. \Cref{sec:problem_formulation} presents the stochastic dynamic programming formulation. \Cref{sec:approach} introduces the prediction-aware adaptive rollout framework and the historical-data-driven fleet-sizing procedure. \Cref{sec:case_study} presents the case study and empirical results. Finally, \Cref{sec:conclusion} concludes the paper and discusses limitations and future directions.

\section{Related Work}
\label{sec:related_work}

This work lies at the intersection of heterogeneous multi-robot task allocation, online scheduling under uncertain demand, request-sequence prediction, and robust planning under forecast mismatch. Prior work has developed expressive models for heterogeneous robot coordination, scalable online allocation mechanisms, predictive models for irregular event sequences, and robustness techniques for model-based decision-making. However, these components are often studied separately. Many online allocation methods react to observed requests but do not explicitly reason about the downstream opportunity cost of current commitments. Prediction-aware methods can anticipate future demand, but often assume that forecasts remain reliable during deployment. Robust planning methods account for uncertainty, but typically do not specify how concrete sampled future requests should be weighted inside an online heterogeneous scheduling policy. Our work addresses this gap by integrating future-request prediction into a rollout-based online planner, adapting the influence of predicted requests using recent forecast errors, and supporting the planner with a historical-data-driven procedure for selecting an initial heterogeneous fleet composition.

\subsection{Heterogeneous Online Task Allocation, Scheduling, and Fleet Composition}

Multi-robot task allocation has been studied through combinatorial optimization, task-and-motion planning, market-based coordination, decentralized algorithms, learning-based policies, and uncertainty-aware planning. These approaches differ in how they model robot heterogeneity, temporal constraints, routing constraints, and online task arrivals. The setting considered in this paper combines several of these difficulties: robots have type-dependent capabilities and travel times, requests have service windows and ordered endpoint sequences, and new real-time requests arrive during execution. The resulting online scheduling problem requires decisions that are computationally tractable, compatible with heterogeneous robot capabilities, and sensitive to future fleet availability.

A first line of work develops expressive models for heterogeneous allocation, scheduling, and integrated task-and-motion planning. Complexity-theoretic results show that multi-robot task allocation is difficult in general \cite{Aziz2021}, and dynamic variants remain challenging when tasks must be inserted, deleted, or rescheduled under heterogeneous capabilities and temporal constraints \cite{Neville2023,Bischoff2024}. Integrated task-and-motion planning methods further couple assignment with sequencing, routing, energy constraints, coalition formation, or learned subteam performance \cite{Chen2021,Neville2021AnIA,Neville2023,Notomista2022,calvo2024,Miloradovic2023,Aswale2023,Gosrich2022MultiRobotCA,Banfi2022}. These methods provide high modeling fidelity, but repeatedly solving detailed coupled planning problems can be impractical in service settings where requests arrive continuously and schedules must be revised frequently.

A second line of work emphasizes scalability and responsiveness through heuristic, auction-based, market-based, decentralized, or metaheuristic mechanisms \cite{Verma2024,Li2021,Zhao2022,Das2015,Bai2024Cluster,Bai2024Impact,Ferreira2022,Wei2020,Kalempa2021}. These methods can often respond quickly to newly observed tasks without solving a large centralized optimization problem. However, purely reactive allocation can be myopic when current assignments affect future service quality. For example, assigning a scarce compatible robot to a low-urgency request may prevent that robot from serving a more urgent request that is likely to arrive later. Our framework addresses this limitation by using sampled future request scenarios to estimate the downstream opportunity cost of current commitments, while still restricting executable assignments to requests that have actually entered the system.

Learning-based allocation methods provide another route to fast online decision-making. Recent work has explored graph neural network schedulers \cite{Wang2022}, imitation-learning-based collaborative schedulers \cite{Gao2023}, reinforcement learning for cooperative task allocation \cite{Park2022RL,shida2024,zhang2025graph,Dai2025}, and LLM-based planning or coordination frameworks \cite{Chen2024LLM,liu2024coherent,Kannan2024,gupta2025,Yang2025}. These approaches can amortize computation through offline training and produce rapid decisions during deployment. However, learned allocation policies can degrade when request arrival patterns, task frequencies, robot availability, travel conditions, or operating policies shift away from the training distribution. In contrast, our approach does not require retraining the allocation policy when prediction quality changes. Instead, it adjusts the planning objective online by reducing the influence of predicted future requests when recent observations indicate forecast mismatch.

Several works incorporate uncertainty directly into heterogeneous allocation, including robust schedules under uncertain robot capabilities \cite{Fu2023}, stochastic trait models with probabilistic task satisfaction \cite{Park2023}, hindsight optimization for uncertain task outcomes \cite{Dhanaraj2024}, and proactive allocation under spatiotemporal uncertainty \cite{street2024right}. These methods show the importance of modeling uncertainty in robot capabilities, task outcomes, and future demand. Our focus is complementary: we consider uncertainty in the future request stream and, more specifically, in the reliability of the predictive distribution used by the online planner. While prior uncertainty-aware allocation methods often model uncertainty in the planning problem itself, they do not typically adapt the influence of predicted future requests based on forecast errors observed during deployment. Our method addresses this gap by estimating prediction reliability online and using it to reweight predicted demand during rollout-based replanning.

The effectiveness of an online assignment policy also depends on the initial heterogeneous robot composition. In heterogeneous service systems, adding robots does not necessarily increase capacity for all requests: a fleet may still fail if it lacks robots compatible with bottleneck request types. Many allocation and scheduling approaches assume that the robot team is fixed before planning begins, while related work on team formation, coalition selection, and subteam performance typically decides which available robots should cooperate on tasks during execution \cite{Neville2021AnIA,Neville2023,Gosrich2022MultiRobotCA,Banfi2022}. These problems are distinct from selecting the initial fleet composition before deployment, where the goal is to ensure that the online assignment policy has sufficient compatible capacity for the demand patterns it is likely to encounter. Our work treats initial heterogeneous fleet sizing as a supporting design decision for online scheduling. We develop a historical-data-driven procedure that evaluates candidate fleet compositions on representative operating days and selects a composition for which the assignment routines used in simulation achieve a target empirical feasibility level.

\subsection{Request-Sequence Prediction and Scenario Generation}

Future service demand in online task-allocation problems can be represented as an irregular sequence of marked events. Each event has an arrival time and a mark encoding task-relevant information such as service type, location, priority, workload, or task family. Temporal point processes are well suited to this setting because they model event timing and marks directly, without requiring demand to be aggregated into fixed time bins. This is useful in sparse or bursty service environments, where discretization can obscure timing structure or introduce many artificial zero-count observations \cite{pmlr-v238-augusto-zagatti24a,chang2026deep}. For heterogeneous robot allocation, the mark structure is especially important because downstream feasibility depends not only on when a request arrives, but also on which robot types can serve it and which service locations must be visited.

Recent work has extended temporal point process models beyond classical formulations by combining continuous-time event modeling with neural sequence architectures. Neural marked temporal point processes have been developed for multivariate and set-valued event data \cite{pmlr-v238-chang24a}, invertible intensity models improve likelihood evaluation and sampling for multivariate processes \cite{pmlr-v238-augusto-zagatti24a}, and state-space point processes combine deep state-space models with continuous-time marked event dynamics \cite{chang2026deep}. Transformer-based event models provide another direction by using attention mechanisms to model dependencies across event histories while preserving continuous and discrete event attributes \cite{draxler2026transformers}. These developments are relevant to robot service systems because they enable predictive models to condition on irregular request histories and generate heterogeneous future request samples.

In our framework, temporal point process predictions are not used as fixed schedules or mandatory tasks. Instead, the prediction module generates sampled future request scenarios from an estimated distribution. Each sampled event is converted into the same request representation used by the planner, including task type, service-window parameters, and graph-grounded endpoint sequence. This representation allows predicted future requests to be evaluated using the same compatibility, routing, and timing constraints. However, because deployment demand may differ from the historical data used to train the predictor, the planner must decide how much influence these sampled requests should have on current decisions. Our contribution is therefore not a new temporal point process architecture, but a planning framework that uses such predictions in a confidence-aware manner. As real requests are observed, recent forecast errors are used to update context-dependent prediction weights, and those weights determine how strongly predicted requests affect rollout values.

\subsection{Robust Planning Under Forecast Error and Distribution Shift}

The value of rollout-based lookahead \cite{bertsekas2019reinforcement, bertsekas2020rollout} depends on the quality of the future request scenarios used in simulation. Predictive models are typically trained on historical data, but deployment conditions may change because of shifts in request arrival rates, task mix, robot availability, travel times, patient or customer behavior, or operating policies. In online allocation, this mismatch is especially consequential because forecast-driven decisions affect future robot availability and may continue to shape the schedule even after prediction errors become apparent. If forecast errors persist, repeated replanning can continue to reintroduce misleading predicted demand unless the planner explicitly reassesses forecast reliability.

Prior work addresses related issues through multi-step prediction, self-correction, adaptation, and retraining. Multi-step prediction and self-correcting dynamics models reduce error accumulation by training over longer rollouts or exposing models to their own past prediction errors \cite{Asadi2019,Talvitie2017}. Other methods adapt learned models after detecting mismatch between real and simulated data through feature alignment, instance reweighting, event-triggered updates, or concept-drift pipelines \cite{Shen2023,ji2022when,ZHANG2025}. These approaches can improve the predictive model itself, but they may require sufficient post-shift data and additional computation before the adapted model becomes useful. Our method instead acts directly at the planning layer: when recent forecast errors increase, the rollout objective immediately reduces the influence of predicted future requests from unreliable contexts.

Other work reduces reliance on model predictions when uncertainty or error is high. A common strategy is to shorten the effective model rollout horizon as model accuracy deteriorates \cite{Jiang2015}, as in short-horizon model-based reinforcement learning \cite{Janner2019}, or to use uncertainty-aware masking schemes that only use model rollouts when they are sufficiently reliable \cite{Pan2020}. Related approaches reweight simulated data, Bellman targets, or rewards according to model quality or uncertainty \cite{Yang2024Weight,Yu2020MOPO}, with similar ideas appearing in adaptive learning and inference-time reweighting \cite{liu2026adaptive,lee2025inferencetime}. Our framework shares the principle that unreliable predictions should have less influence, but differs in how this influence is applied. Rather than discarding predictions beyond a fixed horizon or retraining the predictor, we compute context-dependent confidence weights from observed forecast errors and use them to continuously modulate predicted-request costs during rollout.

Distributionally robust optimization and conformal prediction provide more conservative forms of robustness. Distributionally robust optimization optimizes against a family of plausible distributions and has been applied to stochastic control, routing, vehicle balancing, and other risk-sensitive planning problems \cite{mohajerin2018data,Chaouach2022,Yang2021,Miao2021,Zhang2021,Ji2019,Akella2025Risk,robey2022probabilistically}. Conformal prediction provides calibrated prediction sets or intervals and has been used in robotics, planning, control, online prediction, and safety assurance \cite{vincent2024rollouts,Lindemann2023Planning,Lindemann2025,Gibbs2021,Gibbs2024,Chee2024,Binny2025WhoMM,Lin2025,luo2024sample}. These methods provide principled uncertainty quantification, but they do not directly specify how an online multi-robot allocation policy should weight concrete sampled future requests during replanning. Our approach is operational: it converts recent forecast errors into prediction-confidence weights that directly affect the rollout value of candidate scheduling decisions.

Re-optimization provides a complementary mechanism for robustness. Model predictive control and online allocation methods repeatedly replan as new state information becomes available, which helps limit the effect of disturbances and modeling errors \cite{McAllister2024}. However, replanning alone does not necessarily repair assignments that were made when predictions appeared reliable, especially if those assignments have not yet started but continue to occupy robot capacity. Our framework therefore combines adaptive prediction weighting with selective re-optimization of unstarted assignments. When new observations indicate that the current schedule has become undesirable, eligible assigned but unstarted requests are returned to the pending set and reconsidered under the updated state and prediction-confidence weights. This allows the planner to recover from forecast-driven allocation errors while preserving commitments to services that have already begun.

\section{Problem Formulation}
\label{sec:problem_formulation}

We consider a heterogeneous multi-robot task-assignment and scheduling problem over a finite operating horizon. A fixed robot team must serve scheduled requests known before execution and real-time requests revealed sequentially during operation. Each request has a task type, an ordered sequence of service locations, and timing requirements. Robots are heterogeneous in their task compatibility, travel times, operational status, and assigned maintenance stations / home stations. The online planning problem is to update assignments and schedules as new requests arrive so as to minimize service delay while satisfying compatibility, routing, timing, and end-of-horizon return constraints.

Future real-time requests are uncertain, and the deployment request distribution may differ from the historical distribution used to train the prediction model. We therefore formulate the online scheduling problem as a finite-horizon stochastic dynamic program. The robot team composition is treated as fixed in this formulation; the historical-data-driven procedure used to select this composition before deployment is introduced in \Cref{sec:approach}.

\subsection{Operating Horizon and Environment}
\label{subsec:operating_horizon_environment}

We consider one operating day at a time and index the finite horizon by time steps $t=0,1,\ldots,T$, where $T$ is the horizon length. A control selected at decision time $t\in\{0,\ldots,T-1\}$ determines the evolution of the system over the interval $[t,t+1]$.

The physical environment is represented by a directed traversal graph $  \mathcal{G}=\left(\mathcal{V},\mathcal{E}\right)$, where $\mathcal V$ is the finite set of traversable nodes and $\mathcal E\subseteq\mathcal V\times\mathcal V$ is the set of directed traversal edges. Each node $v\in\mathcal V$ corresponds to a collision-free robot configuration, waypoint, or location in the environment. Each edge $(v,v')\in\mathcal E$ corresponds to a physically realizable local trajectory that moves a robot from node $v$ to node $v'$ without collision. Obstacles and inaccessible regions are therefore represented implicitly by the absence of corresponding nodes or edges in $\mathcal G$.

Only a subset of traversal nodes can appear as request service locations. We denote this subset by $\mathcal V^{\mathrm{svc}} \subseteq \mathcal V$. Nodes in $\mathcal V^{\mathrm{svc}}$ correspond to pickup, drop-off, or task-service locations. Nodes in $\mathcal V\setminus\mathcal V^{\mathrm{svc}}$ are traversal nodes used only for motion planning and do not accept requests.

Let $\mathcal Z$ denote the finite set of robot types. Robot types represent different robot capabilities with distinct available actions. For a robot of type $z\in\mathcal Z$, the time it takes to traverse an edge $(v,v')$ is denoted as $\tau_z(v,v')>0$.

\subsection{Robot Team and Task Compatibility}
\label{subsec:robot_team_composition}

The heterogeneous team composition is $\mathbf M
= (M_z)_{z\in\mathcal Z}$, where $M_z$ is the number of robots of type $z$. The total number of robots is $M = \sum_{z\in\mathcal Z} M_z$. 

The robot identifiers are indexed by $\mathcal L = \{1,\ldots,M\}$. Each robot $\ell\in\mathcal L$ has type $z_\ell\in\mathcal Z$. 

The finite set of task types is denoted by $\mathcal K$. Robot-task compatibility is encoded by $\mathcal K(z) \subseteq \mathcal K$, where $k\in\mathcal K(z)$ means that a robot of type $z$ can serve a request of task type $k$.

The environment contains a set of maintenance stations $\mathcal V^{\mathrm{st}} \subseteq \mathcal V$. Each robot $\ell\in\mathcal L$ is assigned a maintenance station $v_\ell^{\mathrm{st}}\in\mathcal V^{\mathrm{st}}$. Robots start the horizon at their assigned maintenance stations and must return to their assigned maintenance stations no later than time $T$.

The online planning problem is defined for a fixed team composition $\mathbf M$.

\subsection{Requests}
\label{subsec:requests}

A request $r_j$ is described by
\begin{equation}
    r_j =
    \left(
        k_j,
        \boldsymbol{\rho}_j,
        t_j^{\mathrm{entry}},
        t_j^{\mathrm{start}},
        t_j^{\mathrm{des}},
        t_j^{\mathrm{comp}}
    \right).
    \label{eq:request_tuple_simplified}
\end{equation}
Here, $k_j\in\mathcal K$ is the task type. Because the completion of a task may involve a robot visiting a sequence of locations in a specific order, we model the ordered sequence of service nodes that must be visited to complete request $r_j$ as
\begin{equation}
    \boldsymbol{\rho}_j
    =
    \left(
        \rho_{j,1},
        \rho_{j,2},
        \ldots,
        \rho_{j,n_j^{\mathrm{svc}}}
    \right),
    \qquad
    \rho_{j,h}\in\mathcal V^{\mathrm{svc}},
    \label{eq:request_endpoint_sequence_simplified}
\end{equation}
The restriction $\rho_{j,h}\in\mathcal V^{\mathrm{svc}}$ means that requests can only originate, terminate, or require service at designated service nodes; other nodes in $\mathcal V$ are used only for traversal.

The time $t_j^{\mathrm{entry}}$ is the time at which request $r_j$ becomes known to the system, $t_j^{\mathrm{start}}$ is the earliest allowable service start time, $t_j^{\mathrm{des}}$ is the desired completion time, and $t_j^{\mathrm{comp}}$ is the latest allowable completion time. We assume
\begin{equation*}
    t_j^{\mathrm{entry}}
    \leq
    t_j^{\mathrm{start}}
    \leq
    t_j^{\mathrm{des}}
    \leq
    t_j^{\mathrm{comp}}
    \leq
    T.
\end{equation*}
Each task type $k\in\mathcal K$ has an execution time $F_k$. Thus, a feasible schedule for request $r_j$ must visit the service nodes in the order specified by $\boldsymbol{\rho}_j$, respect the earliest-start time $t_j^{\mathrm{start}}$, allocate execution time $F_{k_j}$, and complete the request no later than $t_j^{\mathrm{comp}}$.

Requests are divided into scheduled and real-time requests. The scheduled request set is
\begin{equation}
    \mathcal R^{\mathrm{sched}}
    =
    \left\{
        r_j:
        t_j^{\mathrm{entry}}<0,\;
        0\leq t_j^{\mathrm{start}}
        \leq t_j^{\mathrm{des}}
        \leq t_j^{\mathrm{comp}}
        \leq T
    \right\}.
    \label{eq:scheduled_requests_simplified}
\end{equation}
These requests are known before the operating horizon begins. The real-time request set revealed at time $t$ is
\begin{equation}
    \mathcal R_t^{\mathrm{real}}
    =
    \left\{
        r_j:
        t_j^{\mathrm{entry}}=t,\;
        t\leq t_j^{\mathrm{start}}
        \leq t_j^{\mathrm{des}}
        \leq t_j^{\mathrm{comp}}
        \leq T
    \right\}.
    \label{eq:realtime_requests_simplified}
\end{equation}
The set of requests known after the arrivals at time $t$ have been observed is
\begin{equation}
    \mathcal R_t
    =
    \mathcal R^{\mathrm{sched}}
    \cup
    \bigcup_{\tau=0}^{t}
    \mathcal R_\tau^{\mathrm{real}}.
    \label{eq:known_requests_simplified}
\end{equation}

\subsection{Standing Assumptions}
\label{subsec:standing_assumptions}

We impose the following assumptions throughout the paper.

\begin{assumption}[Well-formed environment and request process]
\label{assum:well_formed_environment}
The traversal graph $\mathcal G=(\mathcal V,\mathcal E)$ and the request process are well formed in the following sense:
\begin{enumerate}
\item The scheduled request set $\mathcal R^{\mathrm{sched}}$ is finite, and the real-time request set $\mathcal R_t^{\mathrm{real}}$ entering the system at each decision time $t=0,\ldots,T-1$ is finite.
\item For every ordered pair of service nodes $v,v'\in\mathcal V^{\mathrm{svc}}$ that may appear as request endpoints, there exists a directed path $P(v,v') = \left(v_0, v_1, \ldots, v_K \right)$ on $\mathcal G$ such that $v_0=v$, $v_K=v'$, $(v_i,v_{i+1})\in\mathcal E$ for all $i=0,\ldots,N-1$, and $v_i\notin\mathcal V^{\mathrm{svc}}$ for $i=1,\ldots,N-1$. 
Thus, any two request endpoints can be connected by a traversal path whose internal nodes are not themselves service locations.
\end{enumerate}
\end{assumption}

\begin{assumption}[Reliable robot execution]
\label{assum:reliable_robot_execution}
Robots execute assigned motion and service actions without failures. In particular, if a robot is assigned a feasible schedule, then it follows the corresponding paths on $\mathcal G$, performs the required service actions, and completes them at the planned times. Thus, uncertainty in the model enters through the real-time request process rather than through robot failures, localization failures, or stochastic task execution.
\end{assumption}

\Cref{assum:well_formed_environment} ensures that the request process is finite over the planning horizon and that service locations are not unavoidable transit bottlenecks in the traversal graph. This allows a motion planner to connect request endpoints without requiring robots to pass through other service nodes as intermediate waypoints. \Cref{assum:reliable_robot_execution} focuses the stochastic component of the model on future request arrivals, rather than on failures in robot motion, localization, or task execution.

\subsection{Stochastic Request Process and Distribution Shift}
\label{subsec:stochastic_request_process}

Let $\omega_t = \mathcal R_t^{\mathrm{real}}$
denote the random real-time requests revealed at decision time $t$. For notational convenience, let $\omega_T=\emptyset$, so that no new requests enter after the final decision time. The request sequence for an operating day is $\mathcal W=(\omega_0,\omega_1,\ldots,\omega_T)$, where the last element is deterministic and empty.
The true deployment distribution of $\mathcal W$ is denoted by $\mathbb P$. This distribution governs the number of real-time requests, their task types, entry times, service windows, desired completion times, and spatial endpoints.

Historical operating days provide request-sequence samples that are used to train a predictive model. The model induces an estimated distribution $\widehat{\mathbb P}$, from which sampled future request scenarios are generated during rollout. We do not assume that $\widehat{\mathbb P} = \mathbb P$.
The mismatch between $\widehat{\mathbb P}$ and $\mathbb P$ represents the distribution shift considered in this paper.

\subsection{Schedules and Feasible Controls}
\label{subsec:schedules_feasible_controls}

At time $t$, each robot has a current node, operational status, assignment, and planned schedule. Let $y_t^\ell$ denote the planned schedule for robot $\ell$. A schedule specifies the future traversal nodes and service actions of robot $\ell$ from time $t$ until its return to its maintenance station.

For a given state $x_t$, let $\mathcal Y_t^\ell(x_t)$
denote the set of feasible schedules for robot $\ell$. A schedule $y_t^\ell\in\mathcal Y_t^\ell(x_t)$ is feasible if it satisfies the following conditions:
\begin{enumerate}
    \item it starts from the current node of robot $\ell$ at time $t$;
    \item it follows valid directed paths on the traversal graph $\mathcal G$;
    \item it assigns only compatible requests to robot $\ell$, meaning $k_j\in\mathcal K(z_\ell)$;
    \item it visits the service nodes of each assigned request $r_j$ in the order specified by $\boldsymbol{\rho}_j$;
    \item it respects each assigned request's earliest-start time and deadline;
    \item it preserves the already executed portion of the current schedule;
    \item it does not reassign a request to a different robot after service has started; and
    \item it returns robot $\ell$ to its maintenance station $v_\ell^{\mathrm{st}}$ no later than time $T$.
\end{enumerate}

The joint schedule is $\mathbf y_t = \left( y_t^1, y_t^2, \ldots, y_t^M \right)$. 
The corresponding feasible joint schedule set is $\mathcal Y_t(x_t) = \mathcal Y_t^1(x_t) \times \cdots \times \mathcal Y_t^M(x_t)$. 

A control at time $t$ is an updated assignment and schedule, $u_t=(\mathbf a_t^u,\mathbf y_t^u)$, where $a_{j,t}^u\in\mathcal L\cup\{-1,\emptyset\}$ records whether each known request $r_j\in\mathcal R_t$ is assigned to a robot, rejected, or left pending, and $\mathbf y_t^u$ is the resulting team schedule. 
The feasible control set is denoted by $\mathcal U_t(x_t)$. Thus, $u_t\in\mathcal U_t(x_t)$
if the updated assignments and schedules satisfy the compatibility, routing, time-window, no-reassignment, and return-to-station constraints described above.

\subsection{State Dynamics}
\label{subsec:state_dynamics}

The state at time $t$, after observing the real-time requests revealed at time $t$, is $x_t = \left( \mathbf p_t, \boldsymbol\theta_t, \mathbf y_t, \mathbf a_t, \boldsymbol\sigma_t, \mathcal R_t \right).$
Here, $\mathbf p_t$ is the vector of nodes currently occupied by the robots, $\boldsymbol\theta_t$ is the vector of robot operational statuses, $\mathbf y_t$ is the current team schedule, $\mathbf a_t$ is the current request assignment vector, $\boldsymbol\sigma_t$ is the vector of request service statuses, and $\mathcal R_t$ is the set of requests known by time $t$.

For each robot $\ell$ the status $\theta_t^{\ell}$ records whether robot $\ell$ is available, executing an assigned schedule, or returning to its maintenance station.
For each request $r_j\in\mathcal R_t$, the assignment status satisfies $a_{j,t} \in \mathcal L \cup \{-1,\emptyset\}$, where $a_{j,t}=\ell$ means that request $r_j$ is assigned to robot $\ell$, $a_{j,t}=\emptyset$ means that it is known but not currently assigned, and $a_{j,t}=-1$ means that it has been rejected. The service status satisfies
\begin{equation}
    \sigma_{j,t}
    \in
    \{
        \mathrm{pending},
        \mathrm{started},
        \mathrm{completed},
        \mathrm{rejected}
    \}.
    \label{eq:service_status}
\end{equation}

After selecting a feasible control $u_t$, robots execute the first step of their updated schedules. The system then observes the next real-time request set $\omega_{t+1}$, and the state evolves according to $x_{t+1} = f_t \left( x_t, u_t, \omega_{t+1} \right)$. 
For notational convenience, let $\omega_T=\emptyset$. For $t=0,\ldots,T-2$, the transition $x_{t+1}=f_t(x_t,u_t,\omega_{t+1})$ incorporates the real-time requests revealed at the next decision time. At the terminal transition, $x_T=f_{T-1}(x_{T-1},u_{T-1},\omega_T)$ with $\omega_T=\emptyset$, so no new requests enter after the final decision time.

\subsection{Service Cost and Planning Objective}
\label{subsec:service_cost_planning_objective}

The planner evaluates assignments according to the service delay and feasibility of the resulting schedules. For a known request $r_j\in\mathcal R_t$ that is assigned to a feasible schedule, let $\widehat C_{j,t}(x_t)$ denote its planned completion time under the current assignment and schedule. The planned delay of request $r_j$ is $h_j(x_t) = \max \left\{ 0, \widehat C_{j,t}(x_t)-t_j^{\mathrm{des}} \right\}$. 

A known request is said to be still serviceable at state $x_t$ if there exists at least one compatible robot schedule, consistent with the constraints in $\mathcal Y_t(x_t)$, that can complete the request no later than $t_j^{\mathrm{comp}}$. Pending serviceable requests are not assigned the infeasibility penalty, since a wait action may intentionally leave such requests unassigned to preserve future capacity.

To compare candidate schedules during online planning, we use a finite penalized request cost that distinguishes between assigned requests, pending requests that can still be served, and requests that have been rejected or have become infeasible. Let $\Phi<\infty$ be a large penalty for rejected or infeasible requests. Define
\begin{equation}
    H_j(x_t)
    =
    \begin{cases}
        h_j(x_t),
        & \text{if } r_j \text{ is assigned to a feasible}\\
        & \text{ schedule or completed}, \\
        0,
        & \text{if } r_j \text{ is pending and can still be}\\
        & \text{ feasibly served}, \\
        \Phi,
        & \text{if } r_j \text{ is rejected or can no longer} \\
        & \text{ be feasibly served}.
    \end{cases}
    \label{eq:request_cost}
\end{equation}
Pending serviceable requests incur no immediate cost, but they remain in the known request set and are evaluated in subsequent decision times. Any request that is not completed or assigned to a feasible schedule by the terminal time is treated as no longer serviceable and receives the penalty $\Phi$.

The aggregate schedule score is
\begin{equation}
    H(x_t)
    =
    \sum_{r_j\in\mathcal R_t}
    H_j(x_t),
    \label{eq:aggregate_schedule_score}
\end{equation}
with the convention that $H(x_t)=0$ if no request has entered the system. The penalty $\Phi$ is chosen large enough to discourage rejection or loss of serviceability, while pending serviceable requests incur no immediate penalty. Thus, a wait action may intentionally leave a request unassigned when preserving capacity is valuable, but the request must eventually be assigned to a feasible schedule or incur the penalty if it is rejected or becomes no longer serviceable.

The one-step cost is defined as the change in aggregate schedule score: $g_t(x_t,u_t,\omega_{t+1}) = H(x_{t+1}) - H(x_t)$, where $x_{t+1} = f_t(x_t,u_t,\omega_{t+1})$. This incremental cost measures how the selected assignment and schedule update changes planned service quality after the next request arrivals are incorporated.

A policy is a sequence $\pi = \left\{ \mu_0, \mu_1, \ldots, \mu_{T-1} \right\}$, where each decision rule maps the current state to a feasible control: $\mu_t(x_t) \in \mathcal U_t(x_t)$. 

The finite-horizon cost-to-go of policy $\pi$ from state $x_t$ is
\begin{equation}
    J_t^{\pi}(x_t)
    =
    \mathbb E_{\mathbb P}
    \left[
        \sum_{\tau=t}^{T-1}
        g_\tau
        \left(
            x_\tau,
            \mu_\tau(x_\tau),
            \omega_{\tau+1}
        \right)
        \,\middle|\,
        x_t
    \right],
    \label{eq:policy_cost_to_go}
\end{equation}
where the expectation is taken over future real-time request arrivals governed by the deployment distribution $\mathbb P$. Because $g_t$ is defined as the change in aggregate schedule score, the finite-horizon objective evaluates the expected net change in planned service quality over the horizon.
Equivalently, up to the fixed baseline $H(x_t)$, the objective minimizes the expected terminal aggregate schedule score.

The ideal planning objective is to find an admissible policy that minimizes this expected cost-to-go:
\begin{equation}
    J_t^{\pi^*}(x_t)
    =
    \min_{\pi\in\Pi}
    J_t^{\pi}(x_t),
    \label{eq:optimal_policy}
\end{equation}
where $\Pi$ is the set of admissible policies. 

The stochastic dynamic program in \eqref{eq:optimal_policy} provides a formal description of the online planning objective, but it cannot be solved exactly in the setting considered here. The control space grows combinatorially because each decision requires selecting assignments, insertion positions, service times, and routes for heterogeneous robots. In addition, evaluating a candidate control requires reasoning about future request arrivals whose distribution is only estimated from historical data and may change during deployment. These difficulties motivate the prediction-aware adaptive rollout framework developed in \Cref{sec:approach}, which approximates the feasible control set, estimates downstream cost through efficient base-policy simulation, adapts the influence of predicted requests using recent forecast errors, and re-optimizes eligible unstarted assignments when the current schedule becomes undesirable.

\section{Our Approach}
\label{sec:approach}

We now develop the prediction-aware adaptive rollout framework motivated by the stochastic dynamic program in \Cref{sec:problem_formulation}. The framework defines an online policy $\tilde{\pi}={\tilde{\mu}_0,\tilde{\mu}_1,\ldots,\tilde{\mu}_{T-1}}$, where each decision rule selects a feasible control from a tractable candidate set, $\tilde{\mu}_t(x_t)\in\widetilde{\mathcal U}_t(x_t),$ where $\widetilde{\mathcal U}_t(x_t)\subseteq\mathcal U_t(x_t)$. The candidate set contains schedule modifications for requests that have actually entered the system. Predicted future requests are not eligible for immediate assignment; they are used only inside rollout simulations to estimate the downstream cost of current decisions.

\subsection{Overview of the Prediction-Aware Adaptive Rollout Framework}
\label{subsec:approach_overview}

At each decision time, the online policy combines computationally tractable candidate generation, sampled future request scenarios, interaction aware base-policy simulation, adaptive prediction confidence weights, and selective re-optimization. After newly observed requests have been incorporated into the state, the planner may return eligible assigned but unstarted requests to the pending set when the current schedule has become undesirable. It then generates a small set of feasible schedule modifications for currently pending requests. Each candidate is evaluated by simulating future decisions over sampled request scenarios from the estimated distribution $\widehat{\mathbb P}$, with predicted requests weighted according to recent forecast accuracy. The selected control updates the real robot schedules, while predicted requests remain hypothetical and influence the decision only through their weighted contribution to simulated future cost.

The rollout policy is implemented using a one-robot-at-a-time decision rule. Instead of optimizing over the full joint control set $\mathcal U_t(x_t)$, the planner processes robots sequentially. When a robot is considered, controls already selected for earlier robots are held fixed, a robot-local candidate set is generated, and each local candidate is evaluated by completing the remaining simulated decisions with an efficient base policy. This decomposition replaces one large joint scheduling problem with a sequence of smaller robot-level rollout decisions, making online evaluation practical while still accounting for interactions through the simulated state, service-node reservations, and future-cost base policy.

The online rollout policy is supported by a pre-deployment fleet-sizing procedure. This procedure uses historical request sequences to select a heterogeneous robot composition for which the assignment routines used inside rollout achieve a target empirical feasibility level on representative operating days. The resulting composition is then treated as fixed during online deployment, as assumed in \Cref{sec:problem_formulation}.

The remainder of this section presents the components of the framework. \Cref{subsec:oat_prediction_aware_rollout} gives the one-robot-at-a-time rollout decision rule. \Cref{subsec:candidate_control_generation} describes candidate-control generation and wait actions. \Cref{subsec:future_request_generation} describes future-request scenario generation. \Cref{subsec:greedy_base_policy} introduces the interaction-aware base policy used for future-cost estimation. \Cref{subsec:adaptive_prediction_confidence} presents adaptive prediction-confidence weighting. \Cref{subsec:reoptimization} describes selective re-optimization of unstarted assignments. Finally, \Cref{subsec:team_sizing} presents the supporting historical fleet-sizing procedure.

\subsection{One-Robot-at-a-Time Rollout Decision Rule}
\label{subsec:oat_prediction_aware_rollout}

We first define the online rollout decision rule. The remaining subsections describe the components used by this rule: candidate-control generation, future-request scenario generation, the interaction-aware base policy, adaptive prediction-confidence weights, selective re-optimization, and the supporting fleet-sizing procedure.

At decision time $t$, the planner has observed state $x_t$ and must select a feasible control from a tractable subset of the full feasible control set. Directly optimizing over the joint candidate set for all robots is expensive because a joint control specifies assignments, insertion positions, service times, and routes for the entire team. We therefore use a one-robot-at-a-time rollout scheme, following the one-agent-at-a-time rollout principle in \cite{bertsekas2019reinforcement, bertsekas2020rollout}. This replaces one large joint optimization with a sequence of smaller robot-local optimizations, while using simulation to account for the effect of each local decision on the rest of the team.

Let $\mathcal O_t = (\ell_1,\ell_2,\ldots,\ell_M)$
denote the order in which robots are processed at decision time $t$. This order may be fixed or chosen from the current state, for example by prioritizing robots that become available earlier. When robot $\ell_i$ is considered, the local controls already selected for robots $\ell_1,\ldots,\ell_{i-1}$ are held fixed. The planner then generates a robot-local candidate set $\widetilde{\mathcal U}_t^{\ell_i} \left( x_t; \widetilde u_t^{\ell_1}, \ldots, \widetilde u_t^{\ell_{i-1}} \right)$, where each candidate modifies only the schedule of robot $\ell_i$ and only uses requests that have entered the system by time $t$. This set includes feasible assignment actions and a wait action, as described in \Cref{subsec:candidate_control_generation}. Predicted future requests are not included in the immediate candidate set.

For a candidate local control 
\begin{align*}
u_t^{\ell_i}\in \widetilde{\mathcal U}_t^{\ell_i}\left( x_t; \widetilde u_t^{\ell_1}, \ldots, \widetilde u_t^{\ell{i-1}} \right),
\end{align*}
we construct a completed joint control by fixing the controls already selected for earlier robots, applying $u_t^{\ell_i}$ to robot $\ell_i$, and using the base policy to complete the simulated decisions for robots that have not yet been processed:
\begin{align*}
    u_t^{\mathrm{comp},i}
    \left(
    u_t^{\ell_i}
    \right)
    =
    \operatorname{Complete}_{\bar\pi}
    \left(
    x_t;
    \widetilde u_t^{\ell_1},
    \ldots,
    \widetilde u_t^{\ell_{i-1}},
    u_t^{\ell_i}
    \right).
\end{align*}
Here, $\bar\pi$ is the interaction-aware base policy introduced in \Cref{subsec:greedy_base_policy}. The completion step is used only to evaluate candidate local controls inside rollout; the executable control applied to the real system is obtained after all robots have been processed.

The candidate $u_t^{\ell_i}$ is evaluated using sampled future request scenarios. Let $\widehat{\mathcal W}_t^s = \left( \widehat{\omega}_{t+1}^s, \widehat{\omega}_{t+2}^s, \ldots, \widehat{\omega}_{\bar T}^s \right)$ be the $s$-th sampled scenario generated from the estimated distribution $\widehat{\mathbb P}$, where $\bar T=\min\{t+D+1,T\}$ and $D$ denotes the rollout depth. Starting from $x_t$, the simulator applies the completed joint control $u_t^{\mathrm{comp},i}(u_t^{\ell_i})$, incorporates the sampled future requests in $\widehat{\mathcal W}_t^s$, and then applies the base policy $\bar\pi$ until the truncated horizon $\bar T$. Let $x_{\bar T}^{s,i,u}$ denote the resulting simulated state.

Because pending serviceable requests have zero immediate cost in the formulation of $H_j$, the rollout value is not computed by scoring an arbitrary partially unresolved state. Instead, before evaluating the terminal score, the base policy is used to resolve pending requests that remain relevant within the truncated rollout horizon. Let
$\operatorname{Resolve}_{\bar\pi}(x)$ denote the simulated state obtained by applying the base policy to the pending requests in state $x$, assigning them if feasible and marking them rejected or no longer serviceable if they cannot be completed before their deadlines. We define the resolved weighted terminal score
\begin{equation}
    H_{\mathrm{res}}^{\lambda}(x)
    =
    H^{\lambda}
    \left(
    \operatorname{Resolve}_{\bar\pi}(x)
    \right),
    \label{eq:resolved_weighted_terminal_score}
\end{equation}
where $H^{\lambda}$ is the weighted aggregate schedule score defined in \Cref{subsec:adaptive_prediction_confidence}. Observed and scheduled requests receive unit weight, while predicted requests are weighted according to the current prediction-confidence values. This resolved score ensures that a request left pending by a wait action is not treated as cost-free indefinitely: during simulation it is either assigned and contributes its planned delay, remains available for assignment at a later simulated decision time, or is eventually marked rejected or no longer serviceable and receives the penalty $\Phi$.

The prediction-aware rollout value of candidate $u_t^{\ell_i}$ is estimated by
\begin{equation}
    \widehat Q_t^{\lambda,i}
    \left(
    x_t,
    u_t^{\ell_i}
    \right)
    =
    \frac{1}{S}
    \sum_{s=1}^{S}
    H_{\mathrm{res}}^{\lambda}
    \left(
    x_{\bar T}^{s,i,u}
    \right).
    \label{eq:oat_weighted_rollout_value}
\end{equation}
Because the one-step cost in \Cref{subsec:service_cost_planning_objective} is defined as a change in aggregate schedule score, this terminal-score form is equivalent, up to a fixed baseline, to summing weighted incremental costs over the truncated rollout horizon.

The selected local control for robot $\ell_i$ is 
\begin{equation}
    \widetilde u_t^{\ell_i}
    \in
    \argmin_{
    v_t^{\ell_i}
    \in
    \widetilde{\mathcal U}_t^{\ell_i}
    \left(
    x_t;
    \widetilde u_t^{\ell_1},
    \ldots,
    \widetilde u_t^{\ell{i-1}}
    \right)
    }
    \widehat Q_t^{\lambda,i}
    \left(
    x_t,
    v_t^{\ell_i}
    \right).
    \label{eq:oat_rollout_decision}
\end{equation}

After the selected control is fixed, the planner proceeds to the next robot in the order $\mathcal O_t$. Once all robots have been processed, the one-robot-at-a-time rollout control is $\widetilde u_t =  \left( \widetilde u_t^{\ell_1}, \widetilde u_t^{\ell_2}, \ldots, \widetilde u_t^{\ell_M} \right)$. This decision rule is prediction-aware because candidate controls are evaluated using sampled future request scenarios. It is adaptive because the contribution of predicted requests to $H^{\lambda}$ depends on online confidence weights computed from recent forecast errors. It avoids premature commitment because predicted requests are never included in the immediate candidate sets; they influence the decision only through their effect on simulated downstream schedule quality.

\subsection{Candidate-Control Generation and Wait Actions}
\label{subsec:candidate_control_generation}

The one-robot-at-a-time rollout rule in \Cref{subsec:oat_prediction_aware_rollout} requires a finite robot-local candidate set. This subsection describes how that set is constructed. The goal is not to enumerate all feasible controls in $\mathcal U_t(x_t)$, but to generate a small set of feasible and promising schedule modifications for currently known requests. Predicted future requests are deliberately excluded from this immediate candidate set; they influence candidate selection only later, through rollout simulation.

At decision time $t$, let $\mathcal D_t$ denote the set of known requests that are eligible for assignment by the candidate generator. After any selective re-optimization step has been applied, we let
\begin{equation}
    \mathcal D_t
    =
    \left\{
    r_j\in\mathcal R_t:
    \sigma_{j,t}=\mathrm{pending},\;
    a_{j,t}=\emptyset
    \right\}.
    \label{eq:pending_request_set_approach}
\end{equation}
Thus, $\mathcal D_t$ contains requests that are known, not rejected, not completed, and not currently fixed in an executing service. Requests whose service has already started are excluded because feasible controls must preserve started assignments. Assigned but unstarted requests enter $\mathcal D_t$ only if the re-optimization mechanism in \Cref{subsec:reoptimization} explicitly releases them back to the pending set.

For each request $r_j\in\mathcal D_t$, the compatible robot set is
\begin{equation}
    \mathcal L_j
    =
    \left\{
    \ell\in\mathcal L:
    k_j\in\mathcal K(z_\ell)
    \right\}.
    \label{eq:compatible_robot_set_approach}
\end{equation}
When robot $\ell$ is processed by the one-robot-at-a-time rollout rule, only requests for which $\ell$ is a compatible robot (i.e. $\ell\in\mathcal L_j$) are considered as assignment candidates for that robot.

A robot-local assignment action assigns a request $r_j\in\mathcal D_t$ to a compatible robot $\ell\in\mathcal L_j$ and appends the service-node sequence $\boldsymbol{\rho}_j$ at the end of the remaining schedule of robot $\ell$. We denote such an action by $a=(r_j,\ell)$. The action is retained only if the resulting schedule is feasible with respect to the constraints in \Cref{subsec:schedules_feasible_controls}: it must follow valid directed paths on $\mathcal G$, visit the service nodes in the order specified by $\boldsymbol{\rho}_j$, satisfy release-time and deadline constraints, preserve the already executed portion of the schedule, and return the robot to its maintenance station by time $T$.

For each action $a=(r_j,\ell)$, we compute two heuristic values used only to rank candidate actions before rollout evaluation. First, we compute a latest feasible service-start time. Let $F_{\ell, j}^{\mathrm{travel}}$ denote the estimated time required for robot $\ell$ to traverse the service-node sequence $\boldsymbol{\rho}_j$, ignoring congestion and conflicts with other robots, but using the type-dependent graph travel-time estimates. The latest feasible service-start time is
\begin{equation}
F_{\ell,j}^{\mathrm{start}}
=
t_j^{\mathrm{comp}}
-
F_{\ell, j}^{\mathrm{travel}}
-
F_{k_j}.
\label{eq:candidate_latest_start_time}
\end{equation}
This value estimates how urgent the request is: smaller values indicate less remaining scheduling flexibility.

Second, we compute a robot-specific heuristic completion delay. Let $F_{\ell, j}^{\mathrm{reach}}$ denote the estimated time at which robot $\ell$ reaches the first service node of $r_j$ if the request is appended to its remaining schedule.  The corresponding heuristic completion time is
\begin{equation}
\widehat C_{\ell, j, t}
=
\max
\left\{
F_{\ell, j}^{\mathrm{reach}},
t_j^{\mathrm{start}}
\right\}
+
F_{\ell j}^{\mathrm{travel}}
+
F_{k_j}.
\label{eq:candidate_heuristic_completion_time}
\end{equation}
If $\widehat C_{\ell, j, t}>t_j^{\mathrm{comp}}$, the action is discarded. Otherwise, the heuristic delay is
\begin{equation}
F_{\ell, j, t}^{\mathrm{delay}}
=
\max
\left\{
0,
\widehat C_{\ell, j, t}
-
t_j^{\mathrm{des}}
\right\}.
\label{eq:candidate_heuristic_delay}
\end{equation}
Candidate assignment actions are ranked lexicographically by urgency and estimated delay:
\begin{equation}
\operatorname{rank}_t(\ell,r_j)
=
\left(
F_{\ell, j}^{\mathrm{start}},
F_{\ell, j, t}^{\mathrm{delay}}
\right).
\label{eq:candidate_ranking_key}
\end{equation}
Actions with earlier latest feasible service-start times are expanded first, and ties are broken by the robot-specific heuristic delay. After sorting by \eqref{eq:candidate_ranking_key}, the generator keeps only the first $L$ feasible assignment actions for robot $\ell$, where $L$ is the candidate limit.

The ranking in \eqref{eq:candidate_ranking_key} is used only for pruning. It is not the rollout objective. After the robot-local candidate set has been generated, each retained assignment action and associated scheduled is evaluated by the rollout estimator in \Cref{subsec:oat_prediction_aware_rollout}, which accounts for sampled future requests, the base policy, and prediction-confidence weights.

The candidate set also includes a wait action, denoted $a_{\ell}^{\mathrm{wait}}$. This action assigns no request to robot $\ell$ during the current local decision. The wait action is always included, even when no feasible assignment action is available, so that the robot-local candidate set is never empty. If the wait action is selected, robot $\ell$ remains unassigned in the real system for the current decision cycle and may be reconsidered at a later decision time.

When a wait action is evaluated inside rollout, robot $\ell$ is temporarily treated as unavailable for the simulated completion of the current local decision. This prevents the base policy used inside the rollout simulation from immediately undoing the wait action by assigning the same robot to another pending request in the same simulated decision. The robot is not removed from the real system; it is simply left available for future decision times. We denote the set of robots that must be treated as unavailable for future cost estimation as $\mathcal{L}^{\mathrm{block}}$.

This procedure reduces the action space used by rollout from the full feasible control set $\mathcal U_t(x_t)$ to a tractable robot-local candidate set $\widetilde{\mathcal U}_t^{\ell_i}$. The retained assignment actions are feasible for the current robot, urgent according to their latest feasible service-start times, and promising according to their heuristic delay estimates. The wait action preserves the planner's ability to defer assignment when the downstream rollout estimate indicates that keeping capacity available is preferable.

\subsection{Future-Request Scenario Generation}
\label{subsec:future_request_generation}

The rollout decision rule in \Cref{subsec:oat_prediction_aware_rollout} evaluates current candidate controls by simulating possible future request arrivals. This subsection describes how those future request scenarios are generated. The scenario generator provides the interface between the learned request-prediction model and the scheduling model in \Cref{sec:problem_formulation}: it samples future events from the estimated distribution $\widehat{\mathbb P}$, converts those events into request tuples with graph-grounded service locations and service windows, and returns sampled disturbance sequences that can be used directly inside rollout simulation.

Historical operating data are used to train temporal point process models for future request generation. Rather than predicting a single global request process over the entire environment, we decompose demand into local request processes. In the hospital case study, each local process corresponds to a patient encounter, but the same construction applies to any setting in which future demand can be associated with local entities, regions, or service contexts. Let $\mathcal P_t$ denote the set of local processes active at decision time $t$. For each process $p\in\mathcal P_t$, let
\begin{equation}
\mathcal H_{p,t}
=
\left(
(\tau_{p,1},\kappa_{p,1}),
\ldots,
(\tau_{p,N_{p,t}},\kappa_{p,N_{p,t}})
\right)
\label{eq:local_tpp_history}
\end{equation}
denote the observed timed-mark history available at time $t$. Here, $\tau_{p,q}$ is the time of previously observed request event $q$ and $\kappa_{p,q}$ is its mark. The mark encodes the task family or request type, together with any discrete attributes used by the trained prediction model. The number of timed marks considered in the history is denoted by $N_{p,t}$.

\begin{figure}
\centering
\includegraphics[width=\linewidth]{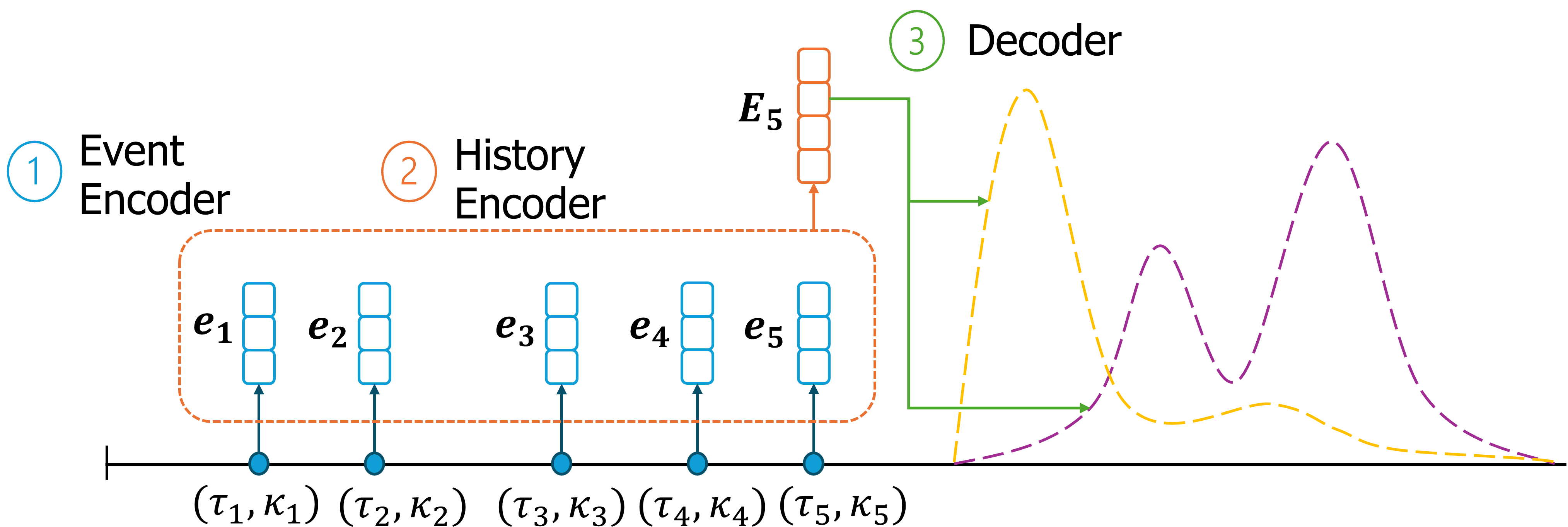}
\caption{\small{Schematic of the neural temporal point process used for future-request generation. Past request events are represented as a sequence of timed marks and encoded into a history-dependent representation. This representation parameterizes conditional distributions over future event times and request marks. Samples from these distributions are then converted into graph-grounded request tuples for rollout simulation.}}
\label{fig:tpp_architecture}
\end{figure}

For each active process $p$, the trained temporal point process defines a conditional distribution over future timed marks. For rollout sample $s$, we draw
\begin{equation}
\left\{
(\widehat\tau_{p,q}^{s},\widehat \kappa_{p,q}^{s})
\right\}_{q\geq 1}
\sim
\widehat{\mathbb P}_{p}
\left(
\cdot
\mid
\mathcal H_{p,t}
\right)
\label{eq:local_tpp_rollout_samples}
\end{equation}
up to the truncated rollout horizon. Figure~\ref{fig:tpp_architecture} illustrates this prediction module. The observed timed-mark prefix is encoded by a neural temporal point process, whose history-dependent representation parameterizes conditional distributions over future event times and marks. Samples from these conditional distributions provide the raw future demand events used by the scenario generator.

The temporal point process samples timed marks, but the rollout simulator requires request tuples of the form defined in \Cref{subsec:requests}. We therefore apply a deterministic request-construction map to each sampled timed mark. Given a sampled event $(\widehat\tau_{p,q}^{s},\widehat \kappa_{p,q}^{s})$, the mark is decoded into a task type $\widehat k_j$ and any task-specific attributes. The sampled time is converted into the desired service time $\widehat t_j^{\mathrm{des}}$, while the entry time, earliest allowable start time, and deadline are constructed from task-specific service-window parameters. We write this conversion abstractly as
\begin{equation}
\widehat r_j
=
\Gamma
\left(
p,
\widehat\tau_{p,q}^{s},
\widehat \kappa_{p,q}^{s}
\right)
=
\left(
\widehat k_j,
\widehat{\boldsymbol{\rho}}_j,
\widehat t_j^{\mathrm{entry}},
\widehat t_j^{\mathrm{start}},
\widehat t_j^{\mathrm{des}},
\widehat t_j^{\mathrm{comp}}
\right).
\label{eq:predicted_request_construction}
\end{equation}
Where $\Gamma$ is a request constructor operator.
The endpoint sequence $\widehat{\boldsymbol{\rho}}_j$ is obtained from the local process context and the task type. In the hospital case study, the process context identifies the patient location at the sampled time. For single-location service requests, this location is the service endpoint. For multi-location requests, such as delivery tasks, the endpoint sequence also includes the relevant supply or pickup location before the destination endpoint. Thus, the learned model predicts irregular timed marks, while the planner converts those marks into graph-grounded service requests using known location and task mappings.

For rollout sample $s$, the generator aggregates all predicted requests into a sampled future disturbance sequence
\begin{equation}
\widehat{\mathcal W}_{t}^{s}
=
\left(
\widehat{\omega}_{t+1}^{s},
\widehat{\omega}_{t+2}^{s},
\ldots,
\widehat{\omega}_{\bar{T}}^{s}
\right),
\qquad
\bar{T}=\min\{t+D+1,T\},
\label{eq:sampled_future_sequence_approach}
\end{equation}
where $D$ is the rollout depth. Each $\widehat{\omega}_{\tau}^{s}$ is a set of predicted requests that enter the simulated system at future decision time $\tau$. Let $\psi(\widehat r_j)$ denote the decision time associated with the predicted entry time $\widehat t_j^{\mathrm{entry}}$. Then $\widehat r_j$ is added to the bucket $\widehat{\omega}_{t'}^{s}$, where $t'=\psi(\widehat r_j)$ and $\psi(\widehat r_j)\in{t+1,\ldots,\bar T}$. Algorithm~\ref{algo:future_request_generator} summarizes the scenario-generation procedure. 

The generator also filters predicted requests that match requests already known to the planner. This prevents the rollout estimator from double-counting scheduled requests or real-time requests already observed in $\mathcal R_t$. A predicted request is treated as matching a known request if it has the same task type, compatible service-node context, and a predicted service time within a tolerance window $\Delta$. Matched predictions are removed before the sampled disturbance sequence is passed to the rollout simulator. 

\begin{algorithm}
\caption{Future-request scenario generation}
\label{algo:future_request_generator}
\begin{algorithmic}[1]
\REQUIRE State $x_t$, active local processes $\mathcal P_t$, local histories $\mathcal H_{p,t}$, local TPP models $\widehat{\mathbb P}_p$, request-construction map $\Gamma$, rollout depth $D$, sample count $S$, match tolerance $\Delta$.
\ENSURE Sampled future request sequences $\widehat{\mathcal W}_t^1,\ldots,\widehat{\mathcal W}_t^S$.
\STATE Set $\bar T\leftarrow\min\{t+D+1,T\}$.
\FOR{$s=1,\ldots,S$}
\STATE Initialize $\widehat{\omega}_{\tau}^{s}\leftarrow\emptyset$ for $\tau=t+1,\ldots,\bar T$.
\FOR{each active local process $p\in\mathcal P_t$}
\STATE Sample future timed marks from $\widehat{\mathbb P}_{p}(\cdot\mid\mathcal H_{p,t})$ up to the truncated horizon.
\FOR{each sampled timed mark $(\widehat\tau_{p,q}^{s},\widehat \kappa_{p,q}^{s})$}
\STATE Construct the predicted request $\widehat r_j=\Gamma(p,\widehat\tau_{p,q}^{s},\widehat \kappa_{p,q}^{s})$.
\IF{$\widehat r_j$ does not match a scheduled or already observed request within tolerance $\Delta$}
\STATE Add $\widehat r_j$ to the future disturbance bucket $\widehat{\omega}_{\psi(\widehat r_j)}^{s}$, if $\psi(\widehat r_j) \in \{t+1,\ldots,\bar T\}$.
\ENDIF
\ENDFOR
\ENDFOR
\STATE Form $\widehat{\mathcal W}_t^s=(\widehat{\omega}_{t+1}^{s},\ldots,\widehat{\omega}_{\bar T}^{s})$.
\ENDFOR
\STATE \textbf{return} $\widehat{\mathcal W}_t^1,\ldots,\widehat{\mathcal W}_t^S$.
\end{algorithmic}
\end{algorithm}
The resulting scenarios have the same request structure as the observed request process in \Cref{sec:problem_formulation}. Consequently, future demand is evaluated inside rollout using the same compatibility, routing, service-window, and scheduling constraints as real requests, while remaining excluded from the immediate candidate-control set.

\subsection{Interaction-Aware Base Policy for Future-Cost Estimation}
\label{subsec:greedy_base_policy}

The one-robot-at-a-time rollout rule evaluates each candidate local control by simulating future system evolution. Because this simulation is repeated across candidate actions, robots, and sampled future request scenarios, the policy used inside rollout must be much cheaper than solving the full online scheduling problem. We therefore use an interaction-aware base policy, denoted $\bar\pi$, to approximate downstream assignment cost after a candidate control has been applied.

The base policy is used in two places. First, in the completion operator $\operatorname{Complete}{\bar\pi}$, it fills in simulated assignments for robots that have not yet been processed by the one-robot-at-a-time rollout rule. Second, in the resolution operator $\operatorname{Resolve}{\bar\pi}$, it attempts to resolve pending requests before the truncated terminal score is evaluated. In both cases, the base policy is used only inside rollout simulation. It is not required to produce the final executable paths applied to the real robots; instead, it provides a fast estimate of how difficult the remaining requests are likely to be after the current candidate decision.

Given a simulated state $\widetilde x$ and a batch of requests $\mathcal R$, the base policy processes requests in priority order. The priority is based on the latest feasible service-start time used in \Cref{subsec:candidate_control_generation}: requests with less remaining scheduling flexibility are considered first. The batch may contain newly sampled future requests, requests left pending by earlier simulated decisions, or requests released for reconsideration in the simulated state. Predicted and observed requests are processed by the same assignment routine; prediction-confidence weights are applied later when the resulting simulated state is scored.

For each queued request $r_j$, the base policy considers compatible robots $\ell\in\mathcal L_j$. Robots that have been temporarily blocked by a wait action in the current rollout evaluation are excluded, so that the base policy cannot immediately undo the candidate wait decision being evaluated. For each remaining compatible robot, the policy estimates the completion time that would result from appending $r_j$ to the robot's simulated schedule. This estimate is calculated using the type-dependent graph travel times through the service-node sequence $\boldsymbol{\rho}_j$, the request service durations, release-time constraints, deadlines, and a service-node reservation table.

\begin{figure}
\centering
\includegraphics[width=\linewidth]{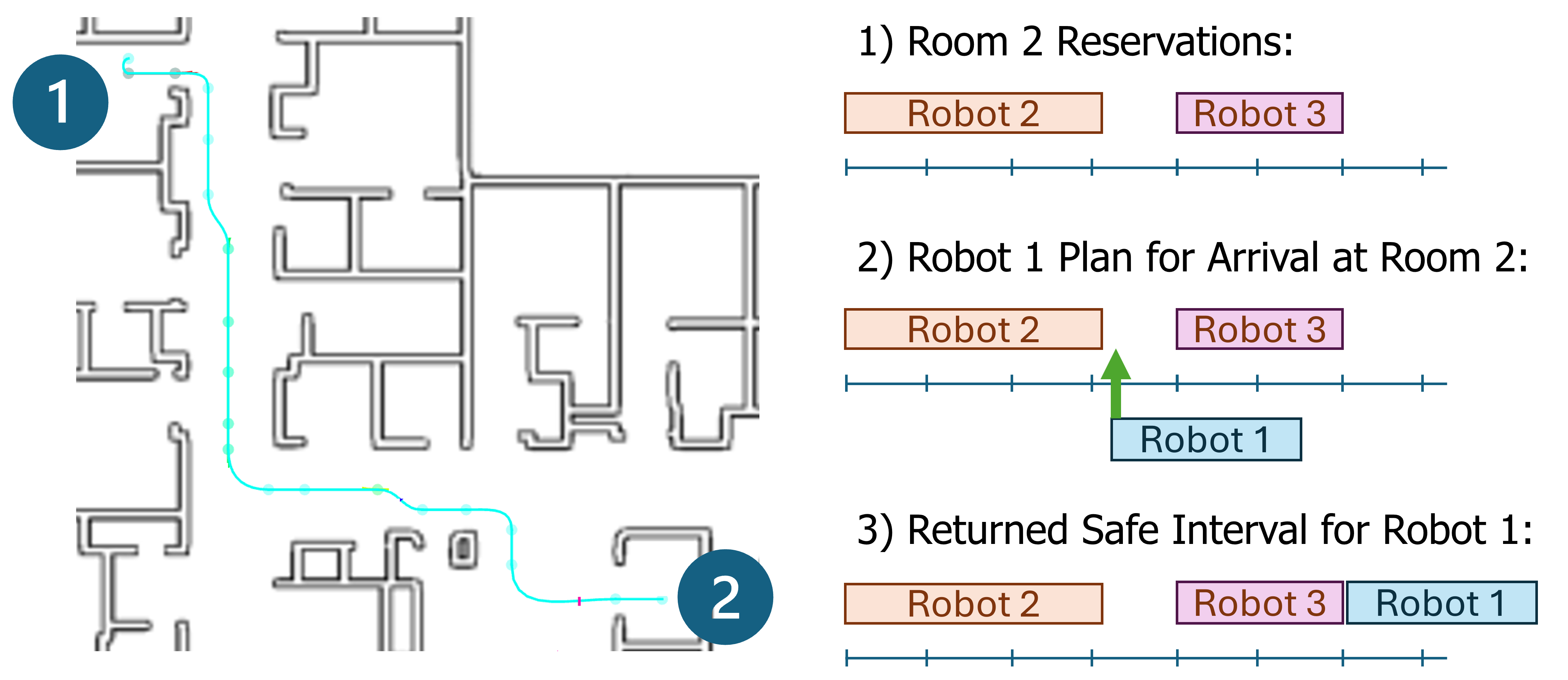}
\caption{\small{Reservation-table check used by the interaction-aware base policy. A candidate service interval is computed from the robot's estimated arrival time and the request execution duration. If the interval overlaps an existing reservation at the same service node, it is shifted to the earliest available interval that satisfies the request's timing constraints. The resulting interval determines the estimated completion time used to compare compatible robot assignments.}}
\label{fig:base_policy_reservations}
\end{figure}

The service-node reservation table is the main interaction-aware component of the base policy. It records time intervals during which simulated robots are expected to occupy task service nodes. When the tentative service interval for a new request overlaps an existing reservation at the same service node, the interval is shifted to the earliest available time that satisfies the request's timing constraints. \Cref{fig:base_policy_reservations} illustrates this reservation-table check. This approximation captures service-location blocking and competition for shared service locations without constructing full collision-free paths for every hypothetical future request.

After estimating feasible completion times for all compatible robots, the request is assigned to the robot with the smallest heuristic delay. If no compatible, unblocked robot can complete the request before its deadline, the request is marked as rejected or no longer serviceable in the simulated state. The resulting simulated assignments, planned completion times, rejected-request indicators, and service-node reservations are then used by the rollout estimator when computing the resolved weighted terminal score. Algorithm~\ref{algo:greedy_base_policy} summarizes the base-policy update. 

\begin{algorithm}
\caption{Interaction-aware base policy used inside rollout}
\label{algo:greedy_base_policy}
\begin{algorithmic}[1]
\REQUIRE Simulated state $\widetilde x$, request batch $\mathcal R$, compatible robot sets ${\mathcal L_j}$, service-node reservation table, blocked robot set $\mathcal L^{\mathrm{block}}$.
\ENSURE Updated simulated state $\widetilde x'$.
\STATE Add requests in $\mathcal R$ to the simulated state if they are not already present.
\STATE Build a priority queue from unresolved requests, ordered by latest feasible service-start time.
\WHILE{the priority queue is not empty}
\STATE Remove the highest-priority request $r_j$.
\STATE Initialize best robot $\ell^* \leftarrow\emptyset$, best delay $c^* \leftarrow+\infty$, and best service intervals $\mathcal I^* \leftarrow\emptyset$.
\FOR{each robot $\ell\in\mathcal L_j\setminus\mathcal L^{\mathrm{block}}$}
\STATE Estimate the earliest feasible service intervals for $r_j$ if appended to robot $\ell$'s simulated schedule.
\STATE Enforce endpoint order, release time, deadline, service duration, and service-node reservation constraints.
\IF{all service nodes of $r_j$ can be served before the deadline}
\STATE Compute the resulting heuristic delay $c$.
\IF{$c<c^*$}
\STATE Set $\ell^*\leftarrow\ell$, $c^*\leftarrow c$, and store the corresponding service intervals in $\mathcal I^*$.
\ENDIF
\ENDIF
\ENDFOR
\IF{$\ell^* = \emptyset$}
\STATE Mark $r_j$ as rejected or no longer serviceable in the simulated state.
\ELSE
\STATE Assign $r_j$ to $\ell^*$, record its planned completion time, and add $\mathcal I^*$ to the reservation table.
\ENDIF
\ENDWHILE
\STATE \textbf{return} the updated simulated state $\widetilde x'$.
\end{algorithmic}
\end{algorithm}
This base policy preserves the constraints that most strongly affect downstream assignment cost: robot-task compatibility, endpoint order, release times, deadlines, service durations, robot availability, and service-location blocking. At the same time, it avoids the cost of constructing full executable schedules for every hypothetical future request in every rollout sample. Full feasibility of the immediate real candidate action is checked before rollout evaluation; the base policy approximation is used only to estimate downstream cost and to rank current candidate controls.

\subsection{Adaptive Prediction Confidence and Weighted Rollout Costs}
\label{subsec:adaptive_prediction_confidence}

The rollout policy uses sampled future requests to estimate downstream cost, but the predictive distribution $\widehat{\mathbb P}$ may differ from the deployment distribution $\mathbb P$. If inaccurate predictions are trusted too strongly, the planner may preserve capacity for requests that do not occur or make current assignments that reflect an outdated demand pattern. We therefore assign confidence weights to predicted requests and update those weights online from recent forecast errors.

The confidence mechanism affects only hypothetical requests used inside rollout simulation. Observed real-time requests and scheduled requests always receive unit weight because they correspond to demand that is known to the planner. Predicted requests receive context-dependent weights. Let $n_j$ denote the prediction context of a predicted request $\widehat r_j$. A context may encode task family, request type, spatial region, floor, time of day, or any other grouping used to estimate prediction reliability. Let $b_j$ denote the confidence-update bin associated with the predicted request $\widehat r_j$. The confidence weight for context $n$ in bin $b$ is denoted by $\lambda_{n,b}\in[\lambda_{\min},1]$, where $\lambda_{\min}\in[0,1]$ is the smallest prediction weight allowed by the planner. A value near one means that predictions in the context are trusted, while smaller values reduce their effect on rollout evaluation.

For a simulated state $x$, let $\mathcal R^{\mathrm{obs}}(x)$ denote the observed or scheduled requests in the state, and let $\widehat{\mathcal R}(x)$ denote the predicted requests present in the simulated state. The weighted aggregate schedule score is
\begin{equation}
H^{\lambda}(x)
=
\sum_{r_j\in\mathcal R^{\mathrm{obs}}(x)}
H_j(x)
+
\sum_{\widehat r_j\in\widehat{\mathcal R}(x)}
\lambda_{n_j,b_j}
H_{\widehat r_j}(x).
\label{eq:weighted_schedule_score_approach}
\end{equation}
This score is used only inside rollout simulation. It does not alter the real system state, the feasibility constraints, or the commitments made to observed requests. In the one-robot-at-a-time rollout rule, $H^{\lambda}$ is used through the resolved terminal score $H_{\mathrm{res}}^{\lambda}$, so that pending predicted requests are resolved by the base policy before their weighted contribution is evaluated.

The weights are updated by comparing recent observations with prediction snapshots generated before those observations were available. Prediction confidence is evaluated on fixed bins of duration $\Delta_\lambda$. Let $\xi$ index the confidence-update bin, and let $\mathcal T_\xi$ denote the short set of bins used for the comparison. In the implementation, $\mathcal T_\xi$includes the target bin and adjacent bins used to tolerate small timing shifts. Let $O_{n,\iota}$ be the observed request count in context $n$ and comparison bin $\iota\in\mathcal T_\xi$, and let $\widehat O_{n,\iota}^s$ be the corresponding predicted count in rollout sample $s$. The update distinguishes two forecast errors: overprediction, where predicted requests fail to occur, and timing error, where predicted demand appears near the predicted time but in an adjacent bin.

The sampled predictions define the mean predicted count
\begin{equation}
\mu_{n,\iota}
=
\frac{1}{S}
\sum_{s=1}^{S}
\widehat O_{n,\iota}^{s}.
\label{eq:prediction_mean_count}
\end{equation}
We also compute empirical surprise scores from the sampled predictive counts. With add-one smoothing,
\begin{align}
\varepsilon_{n,\iota}^{\mathrm{over}}
&=
\frac{
1+\sum_{s=1}^{S}\mathbb{1}_{\{\widehat O_{n,\iota}^{s}\leq O_{n,\iota}\}}
}{S+1}, \\
E_{n,\iota}^{\mathrm{over}}
& =
-\log\max \{\varepsilon_{n,\iota}^{\mathrm{over}},\epsilon \} \\
\varepsilon_{n,\iota}^{\mathrm{under}}
&=
\frac{
1+\sum_{s=1}^{S}\mathbb{1}_{\{\widehat O_{n,\iota}^{s}\geq O_{n,\iota}\}}
}{S+1},
\\
E_{n,\iota}^{\mathrm{under}}
&=
-\log\max \{\varepsilon_{n,\iota}^{\mathrm{under}},\epsilon \},
\label{eq:prediction_surprise_scores}
\end{align}
where $\epsilon>0$ prevents undefined logarithms. The overprediction surprise $E_{n,\iota}^{\mathrm{over}}$ is large when the realized count is unusually low under the predictive samples, while the underprediction surprise $E_{n,\iota}^{\mathrm{under}}$ is large when the realized count is unusually high.

To separate spurious predictions from small temporal shifts, we first compute predicted and observed excess counts,
\begin{equation}
\varrho_{n,\iota}^{+}
=
[\mu_{n,\iota}-O_{n,\iota}]_{+},
\qquad
\varphi_{n,\iota}^{+}
=
[O_{n,\iota}-\mu_{n,\iota}]_{+},
\label{eq:prediction_excess_masses}
\end{equation}
where $[z]_+=\max\{z,0\}$. Predicted excess mass in bin $\iota$ may be matched to observed excess mass in an adjacent bin $\iota'$, with $|\iota-\iota'|=1$. Let $m_{n,\iota,\iota'}$ denote the amount of predicted excess mass in bin $\iota$ matched to observed excess mass in bin $\iota'$. This matched mass is treated as timing error rather than as a completely spurious prediction. The remaining unmatched predicted excess is
\begin{equation}
P_{n,\iota}^{\mathrm{over}}
=
\varrho_{n,\iota}^{+}
-
\sum_{\substack{\iota'\in\mathcal T_\xi:\ |\iota-\iota'|=1}}
m_{n,\iota,\iota'}.
\label{eq:residual_overprediction_mass}
\end{equation}
The instantaneous error scores are normalized by the predicted mass in the evaluated window,
\begin{equation}
B_{n,\xi}
=
\epsilon
+
\sum_{\iota\in\mathcal T_\xi}
\mu_{n,\iota}.
\label{eq:prediction_mass_normalizer}
\end{equation}
The overprediction score is
\begin{equation}
A_{n,\xi}^{\mathrm{over}}
=
\frac{1}{B_{n,\xi}}
\sum_{\iota \in\mathcal T_\xi}
P_{n,\iota}^{\mathrm{over}}
E_{n,\iota}^{\mathrm{over}},
\label{eq:instantaneous_overprediction_score}
\end{equation}
and the timing-error score is
\begin{equation}
A_{n,\xi}^{\mathrm{time}}
=
\frac{1}{B_{n,\xi}}
\sum_{\iota\in\mathcal T_\xi}
\sum_{\substack{\iota'\in\mathcal T_\xi:\ |\iota-\iota'|=1}}
m_{n,\iota,\iota'}
\frac{
E_{n,\iota}^{\mathrm{over}}
+
E_{n,\iota'}^{\mathrm{under}}
}{2}.
\label{eq:instantaneous_timing_score}
\end{equation}

Thus, $A_{n,\xi}^{\mathrm{over}}$ penalizes predicted mass that does not materialize even after adjacent-bin tolerance, while $A_{n,\xi}^{\mathrm{time}}$ penalizes predicted mass that appears shifted to a neighboring bin. Unmatched observed excess corresponds to underprediction. We record this diagnostically, but do not use it to decrease prediction confidence, because newly observed requests enter the real state with unit weight and can be handled directly by the online planner and the re-optimization mechanism.

The instantaneous scores are smoothed over time:
\begin{equation}
S_{n,\xi}^{\mathrm{over}}
=
(1-\alpha_{\mathrm{over}})
S_{n,\xi-1}^{\mathrm{over}}
+
\alpha_{\mathrm{over}}
A_{n,\xi}^{\mathrm{over}},
\label{eq:overprediction_ema}
\end{equation}
and
\begin{equation}
S_{n,\xi}^{\mathrm{time}}
=
(1-\alpha_{\mathrm{time}})
S_{n,\xi-1}^{\mathrm{time}}
+
\alpha_{\mathrm{time}}
A_{n,\xi}^{\mathrm{time}},
\label{eq:timing_error_ema}
\end{equation}
where $\alpha_{\mathrm{over}},\alpha_{\mathrm{time}}\in[0,1]$ control the responsiveness of the update. The confidence weight is then
\begin{equation}
\lambda_{n,\xi} 
=
\lambda_{\min}
+
(1-\lambda_{\min})
\exp
\left(
-\beta_{\mathrm{over}}S_{n,\xi}^{\mathrm{over}}
-\beta_{\mathrm{time}}S_{n,\xi}^{\mathrm{time}}
\right),
\label{eq:prediction_weight_update}
\end{equation}
where $\beta_{\mathrm{over}},\beta_{\mathrm{time}}\geq0$ control how strongly the smoothed errors reduce confidence. Persistent overprediction or timing mismatch therefore lowers the weight assigned to future predicted requests in the same context. In our implementation, we typically choose $\beta_{\mathrm{over}}\geq\beta_{\mathrm{time}}$, because persistent overprediction can cause rollout to reserve capacity for demand that does not occur, whereas small timing shifts may still correspond to nearby real demand. If a prediction snapshot contains no positive predicted mass for a context, the update for that context is skipped and the previous smoothed error states are retained.

For sparse local processes, the implementation uses a hierarchical fallback to avoid overreacting to limited evidence. Let $\lambda_{n,\xi}^{\mathrm{loc}}$ be the local context weight, let $\lambda_{n,\xi}^{\mathrm{fb}}$ be a broader fallback-context weight, and let $c_{n,\xi}$ be the number of positive-prediction windows observed so far for context $n$. The local credibility coefficient is
\begin{equation}
\chi_{n,\xi}
=
\frac{c_{n,\xi}}{c_{n,\xi}+\zeta},
\label{eq:local_credibility_blend}
\end{equation}
where $\zeta\geq0$ is a credibility prior. The effective weight applied to a predicted request is
\begin{equation}
\lambda_{n,\xi}^{\mathrm{eff}}
=
\chi_{n,\xi}\lambda_{n,\xi}^{\mathrm{loc}}
+
(1-\chi_{n,\xi})\lambda_{n,\xi}^{\mathrm{fb}}.
\label{eq:effective_prediction_weight}
\end{equation}
When hierarchical fallback is enabled, the symbol $\lambda_{n,\xi}$ in \eqref{eq:weighted_schedule_score_approach} denotes this effective weight. Thus, new or sparse contexts initially rely more heavily on a broader reliability estimate, while local context weights dominate after sufficient prediction evidence has accumulated. Algorithm~\ref{algo:prediction_weight_update} summarizes the confidence-update procedure. 

\begin{algorithm}
\caption{Online prediction-confidence update}
\label{algo:prediction_weight_update}
\begin{algorithmic}[1]
\REQUIRE Prediction snapshot $\{\widehat O_{n,\iota}^{s}:\iota\in\mathcal T_\xi\}_{s=1}^{S}$, observed counts $\{O_{n,\iota}:\iota\in\mathcal T_\xi\}$, previous states $S_{n,\xi-1}^{\mathrm{over}}$ and $S_{n,\xi-1}^{\mathrm{time}}$, hyperparameters $\alpha_{\mathrm{over}}$, $\alpha_{\mathrm{time}}$, $\beta_{\mathrm{over}}$, $\beta_{\mathrm{time}}$, $\lambda_{\min}$, and $\epsilon$.
\ENSURE Updated confidence weight $\lambda_{n,\xi}$.
\STATE Compute mean predicted counts $\mu_{n,\iota}$ and surprise scores $E_{n,\iota}^{\mathrm{over}}$, $E_{n,\iota}^{\mathrm{under}}$.
\STATE Compute predicted excess $\varrho_{n,\iota}^{+}$ and observed excess $\varphi_{n,\iota}^{+}$.
\STATE Match predicted excess to adjacent-bin observed excess, yielding timing matches $m_{n,\iota,l\iota'}$.
\STATE Compute residual overprediction mass $P_{n,\iota}^{\mathrm{over}}$.
\STATE Compute $A_{n,\xi}^{\mathrm{over}}$ and $A_{n,\xi}^{\mathrm{time}}$.
\STATE Update $S_{n,\xi}^{\mathrm{over}}$ and $S_{n,\xi}^{\mathrm{time}}$ using \eqref{eq:overprediction_ema} and \eqref{eq:timing_error_ema}.
\STATE Compute $\lambda_{n,\xi}$ using \eqref{eq:prediction_weight_update}.
\STATE Apply hierarchical fallback, if enabled, using \eqref{eq:effective_prediction_weight}.
\STATE \textbf{return} the effective confidence weight.
\end{algorithmic}
\end{algorithm}

Adaptive prediction confidence provides prospective correction. It does not change real assignments already made by the planner; rather, it changes how strongly future predicted requests influence subsequent rollout evaluations. Retrospective correction of already assigned but unstarted requests is handled by the selective re-optimization mechanism in \Cref{subsec:reoptimization}.

\subsection{Selective Re-optimization of Unstarted Assignments}
\label{subsec:reoptimization}

Adaptive prediction confidence provides prospective correction: it changes how strongly predicted future requests influence subsequent rollout evaluations. However, changing the prediction weights does not automatically repair assignments that were already made when those predictions appeared reliable. The planner may therefore remain committed to a schedule that is no longer desirable after new requests arrive or after recent observations reveal forecast mismatch. To address this issue, we include a selective re-optimization step for requests that have been assigned but whose service has not yet started.

Let
\begin{equation}
\mathcal D_t^{\mathrm{unstarted}}
=
\left\{
r_j\in\mathcal R_t:
a_{j,t}\in\mathcal L,
\sigma_{j,t}=\mathrm{pending}
\right\}
\label{eq:unstarted_assigned_set_approach}
\end{equation}
denote the set of assigned but unstarted requests at decision time $t$. These requests may be returned to the pending set because changing their assignment does not violate the no-reassignment constraint for started requests. Requests whose service has already begun remain fixed until completion.

To decide when re-optimization is useful, we compare the urgency of newly observed requests with the urgency of assigned but unstarted requests. Let $\underline C_{j,t}$ be a lower bound on the earliest possible completion time of request $r_j$ from state $x_t$, obtained using shortest-path travel-time estimates and the request execution duration. We define the slack index
\begin{equation}
\varsigma_{j,t}
=  t_j^{\mathrm{comp}} - \underline C_{j,t}.
\label{eq:request_slack_index_approach}
\end{equation}
Smaller values of $\varsigma_{j,t}$ indicate less remaining flexibility before the request deadline.
Re-optimization is considered only when both newly observed requests and assigned but unstarted requests are present.The re-optimization trigger is 
\begin{equation}
\min_{r_j\in\mathcal \mathcal R_t^{\mathrm{real}}}
\varsigma_{j,t}
<
\min_{r_i\in\mathcal D_t^{\mathrm{unstarted}}}
\varsigma_{i,t}.
\label{eq:reoptimization_trigger}
\end{equation}
When \eqref{eq:reoptimization_trigger} holds, at least one newly observed request is more urgent than every currently assigned but unstarted request. In this case, the planner releases lower-priority unstarted assignments back to the pending set. In particular, an assigned but unstarted request $r_i\in\mathcal D_t^{\mathrm{unstarted}}$ is eligible for release if its slack is larger than the slack of the most urgent newly observed request: $\varsigma_{i,t} >
\min_{r_j\in\mathcal \mathcal R_t^{\mathrm{real}}}
\varsigma_{j,t}.$

Released requests have their assignment reset to $a_{i,t}=\emptyset$ and are added to $\mathcal D_t$. The affected robot schedules are then repaired by removing the released requests while preserving all started or completed service commitments.

After this release step, the one-robot-at-a-time rollout policy is applied to the updated state. Newly observed requests and released unstarted requests are considered together by the candidate generator, and candidate controls are evaluated using the current prediction-confidence weights. Thus, re-optimization allows the planner to revise eligible commitments when new information makes the previous schedule undesirable.

This step provides retrospective correction. Prediction-confidence weighting reduces the influence of unreliable future predictions in subsequent rollout evaluations, whereas selective re-optimization allows the planner to recover from assignments made before those forecast errors were detected.

\subsection{Supporting Pre-Deployment Fleet Sizing}
\label{subsec:team_sizing}

The online planning problem in \Cref{sec:problem_formulation} is defined for a fixed heterogeneous team composition $\mathbf M=(M_z)_{z\in\mathcal Z}$. The rollout policy developed above assumes that this composition is given during deployment. However, the quality of rollout estimates depends on whether the team has enough capacity of the right robot types to serve the request patterns likely to occur in operation. If the assignment routines used inside rollout frequently reject requests or miss deadlines, then simulated future costs become dominated by infeasibility rather than by meaningful comparisons among current candidate controls. We therefore use a supporting pre-deployment procedure to select a heterogeneous team composition from historical request sequences.

Let $\mathcal H^{\mathrm{day}}$ denote a set of historical operating days. Each day $d\in\mathcal H^{\mathrm{day}}$ provides a realized request sequence $\mathcal W^d = \left( \omega_0^d, \omega_1^d, \ldots, \omega_{T-1}^d \right)$. The goal is to choose a fixed composition $\mathbf M$ for which the assignment routine used in simulation achieves a target empirical feasibility level $\gamma\in(0,1]$ on these representative days. A historical day is considered feasible for composition $\mathbf M$ if all requests are assigned to compatible robots, completed no later than their deadlines, and all robots return to their assigned maintenance stations by time $T$.

Our proposed team-sizing procedure has two phases. The first phase computes day-wise feasible compositions and uses them to initialize the team size. For each historical day $d$, the procedure starts from a small baseline composition and simulates the assignment routine on $\mathcal W^d$. If a request is rejected or violates its service window, the procedure increases the count of a robot type compatible with the failed request and repeats the simulation. Once all requests on day $d$ are served feasibly, the resulting day-wise composition is recorded as $\mathbf M^d =(M_z^d)_{\{z\in\mathcal Z\}}$. These day-wise counts estimate the amount of type-specific capacity required by individual historical operating days.

For each robot type $z$, we form the empirical distribution of the day-wise feasible counts $\{M_z^d:d\in\mathcal H^{\mathrm{day}}\}$. Let
\begin{equation}
\widehat p_z(m)
=
\frac{1}{|\mathcal H^{\mathrm{day}}|}
\sum_{d\in\mathcal H^{\mathrm{day}}}
\mathbb 1_{\{M_z^d\leq m\}}
\label{eq:team_size_empirical_cdf}
\end{equation}
be the empirical distribution function for type $z$, where $\mathbb 1_{\{M_z^d\leq m\}}$ is an indicator variable. The initial team size for type $z$ is chosen as the $\gamma$-quantile of this empirical distribution:
\begin{equation}
M_z^{(0)}
=
\min
\left\{
m\in\mathbb Z_{\geq 0}:
\widehat p_z(m)\geq \gamma
\right\}.
\label{eq:team_size_gamma_percentile}
\end{equation}
The resulting initialization is $\mathbf M^{(0)} = (M_z^{(0)})_{\{z\in\mathcal Z\}}$. 

For example, if $\gamma=0.95$, then $M_z^{(0)}$ is the smallest count such that at least $95\%$ of the historical single-day sizing runs required no more than $M_z^{(0)}$ robots of type $z$. Algorithm~\ref{algo:team_sizing_initialization} summarizes this initialization phase.

\begin{algorithm}
\caption{Histogram-based initialization of the heterogeneous team composition}
\label{algo:team_sizing_initialization}
\begin{algorithmic}[1]
\REQUIRE Historical days $\mathcal H^{\mathrm{day}}$, target level $\gamma$, baseline composition $\mathbf M^{\mathrm{base}}$.
\ENSURE Initial team composition $\mathbf M^{(0)}$.
\FOR{each historical day $d\in\mathcal H^{\mathrm{day}}$}
\STATE Set $\mathbf M^d\leftarrow \mathbf M^{\mathrm{base}}$.
\REPEAT
\STATE Simulate the assignment routine on $\mathcal W^d$ using team $\mathbf M^d$.
\IF{all requests are completed feasibly and robots return by $T$}
\STATE Mark day $d$ feasible under $\mathbf M^d$.
\ELSE
\STATE Select a rejected or deadline-violating request $r_j$.
\STATE Select a robot type $z$ such that $k_j\in\mathcal K(z)$.
\STATE Increase $M_z^d\leftarrow M_z^d+1$.
\ENDIF
\UNTIL{day $d$ is feasible}
\STATE Record $\mathbf M^d=(M_z^d)_{\{z\in\mathcal Z \}}$.
\ENDFOR
\FOR{each robot type $z\in\mathcal Z$}
\STATE Compute $\widehat p_z$ from $\{M_z^d:d\in\mathcal H^{\mathrm{day}}\}$.
\STATE Set $M_z^{(0)}\leftarrow \min \{m\in\mathbb Z_{\geq 0}:\widehat p_z(m)\geq\gamma \}$.
\ENDFOR
\STATE \textbf{return} $\mathbf M^{(0)}$.
\end{algorithmic}
\end{algorithm}

The second phase verifies the joint composition across all historical days. This step is necessary because the percentile initialization treats robot types marginally: choosing a high percentile for each type separately does not guarantee that the combined heterogeneous team achieves the desired feasibility rate when task interactions, service windows, routing, and robot-type substitution are evaluated jointly.

For a candidate composition $\mathbf M$, define the empirical feasibility rate
\begin{equation}
\widehat p_{\mathrm{feas}}(\mathbf M)
=
\frac{1}{|\mathcal H^{\mathrm{day}}|}
\sum_{d\in\mathcal H^{\mathrm{day}}}
\mathbb 1_{
\left\{
\text{day $d$ is feasible under $\mathbf M$}
\right\}}.
\label{eq:empirical_feasibility_rate_approach}
\end{equation}
where $\mathbb 1$ is an indicator variable. The verification phase starts from $\mathbf M^{(0)}$ and repeatedly simulates the historical days. If $\widehat p_{\mathrm{feas}}(\mathbf M)\geq\gamma$, the current composition is accepted. Otherwise, the procedure identifies which robot types are associated with failed requests and increments the type with the largest failure score.

Let $\mathcal R_{\mathrm{fail}}^d(\mathbf M)$ denote the set of requests that are rejected or violate their deadlines on historical day $d$ when using composition $\mathbf M$. The failure score of robot type $z$ is
\begin{equation}
B_z(\mathbf M)
=
\sum_{d\in\mathcal H^{\mathrm{day}}}
\sum_{r_j\in\mathcal R_{\mathrm{fail}}^d(\mathbf M)}
\mathbb 1_{
\left\{
k_j\in\mathcal K(z)
\right\}}.
\label{eq:type_failure_score}
\end{equation}
This score counts how often robots of type $z$ could have served failed requests. The search increases the count of a type with the largest failure score and repeats the historical verification. Algorithm~\ref{algo:team_sizing} summarizes this phase.

\begin{algorithm}
\caption{Historical verification of the heterogeneous team composition}
\label{algo:team_sizing}
\begin{algorithmic}[1]
\REQUIRE Historical days $\mathcal H^{\mathrm{day}}$, target feasibility level $\gamma$, initial composition $\mathbf M^{(0)}$.
\ENSURE Verified team composition $\mathbf M$.
\STATE Set $\mathbf M\leftarrow \mathbf M^{(0)}$.
\REPEAT
\STATE Set $N_{\mathrm{feas}}\leftarrow 0$ and $B_z\leftarrow 0$ for all $z\in\mathcal Z$.
\FOR{each historical day $d\in\mathcal H^{\mathrm{day}}$}
\STATE Simulate the assignment routine on $\mathcal W^d$ using team $\mathbf M$ and base policy $\bar \pi$.
\IF{all requests are completed feasibly and robots return by $T$}
\STATE Set $N_{\mathrm{feas}}\leftarrow N_{\mathrm{feas}}+1$.
\ELSE
\STATE Let $\mathcal R_{\mathrm{fail}}^d(\mathbf M)$ be the rejected or deadline-violating requests.
\FOR{each $r_j\in\mathcal R_{\mathrm{fail}}^d(\mathbf M)$}
\FOR{each type $z\in\mathcal Z$ such that $k_j\in\mathcal K(z)$}
\STATE Set $B_z\leftarrow B_z+1$.
\ENDFOR
\ENDFOR
\ENDIF
\ENDFOR
\STATE Compute $\widehat p_{\mathrm{feas}}(\mathbf M)=N_{\mathrm{feas}}/|\mathcal H^{\mathrm{day}}|$.
\IF{$\widehat p_{\mathrm{feas}}(\mathbf M)<\gamma$}
\STATE Select $z^* \in \argmax_{z\in\mathcal Z}B_z$, using a deterministic tie-breaker.
\STATE Increase $M_{z}\leftarrow M_{z^*}+1$.
\ENDIF
\UNTIL{$\widehat p_{\mathrm{feas}}(\mathbf M)\geq\gamma$}
\STATE \textbf{return} $\mathbf M$.
\end{algorithmic}
\end{algorithm}
This procedure does not prove feasibility under the deployment distribution $\mathbb P$. It selects a fixed heterogeneous team composition for which the assignment routine is empirically feasible on representative historical operating days. The selected composition is then used as the fixed team $\mathbf M$ assumed by the online problem formulation and by the prediction-aware rollout policy.

\section{Case Study: Task Assignment in Hospital Floors}
\label{sec:case_study}

We evaluate the proposed framework on a hospital-floor task-assignment problem constructed from historical service requests on inpatient floors. This setting captures the main features of the formulation in \Cref{sec:problem_formulation}: requests arrive over time, different task families require different robot capabilities, requests have service windows and deadlines, and current assignment decisions affect the robot capacity available for future demand. It also provides a realistic setting in which the predictive request model may be useful on average but unreliable in particular time periods, floors, or task contexts.

The case study has three goals. First, we evaluate whether the temporal point process models used for scenario generation produce useful sampled future request sequences. Second, we evaluate the pre-deployment fleet-sizing procedure from \Cref{subsec:team_sizing}, which selects the fixed heterogeneous team composition used by the online policies. Third, we compare the full prediction-aware adaptive rollout policy against baseline assignment policies and against ablated rollout variants. These comparisons isolate the contribution of sampled future-demand information, adaptive prediction reweighting, and selective re-optimization of unstarted assignments.

The results are presented in the same order as the components of the proposed framework. We first evaluate the temporal point process predictions used to generate future scenarios. We then report the fleet-sizing results that determine the fixed monitoring and delivery robot teams. Next, we compare the full adaptive rollout policy against baseline online assignment policies over held-out floor-days. Finally, we present ablation studies that separate the effects of prediction reweighting and re-optimization, followed by runtime results that assess whether the rollout computation is suitable for online use.

\subsection{Experimental Setup}
\label{subsec:case_study_setup}

\begin{figure}
    \centering
    \includegraphics[width=\linewidth]{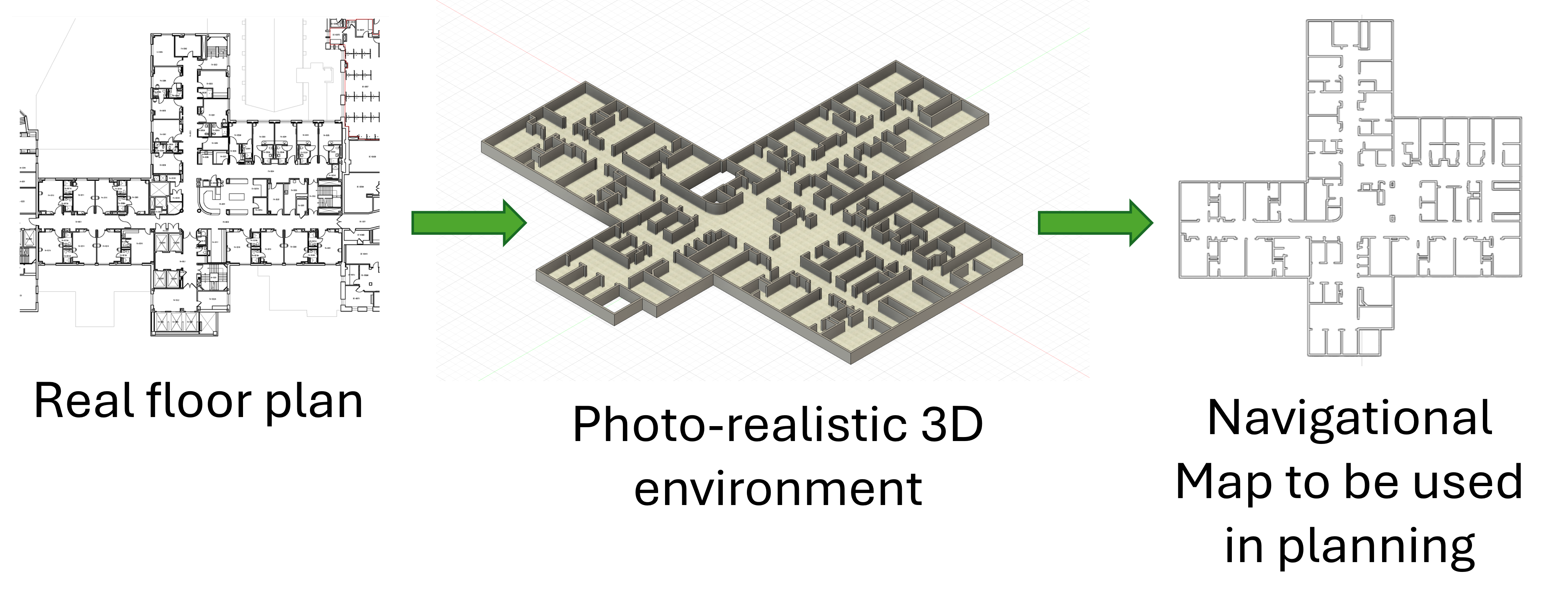}
    \caption{\small{Environment-generation pipeline used in the case study. Starting from an architectural floor plan, we construct a simplified three-dimensional representation of the hospital floor and then extract a two-dimensional navigation map for robot simulation and planning. The resulting map defines the traversable workspace used to build the graph representation, compute travel-time heuristics, and assign service locations for scheduled, observed, and predicted requests.}}
    \label{fig:env_generation}
\end{figure}

\begin{figure}
    \centering 
    \includegraphics[width=\linewidth]{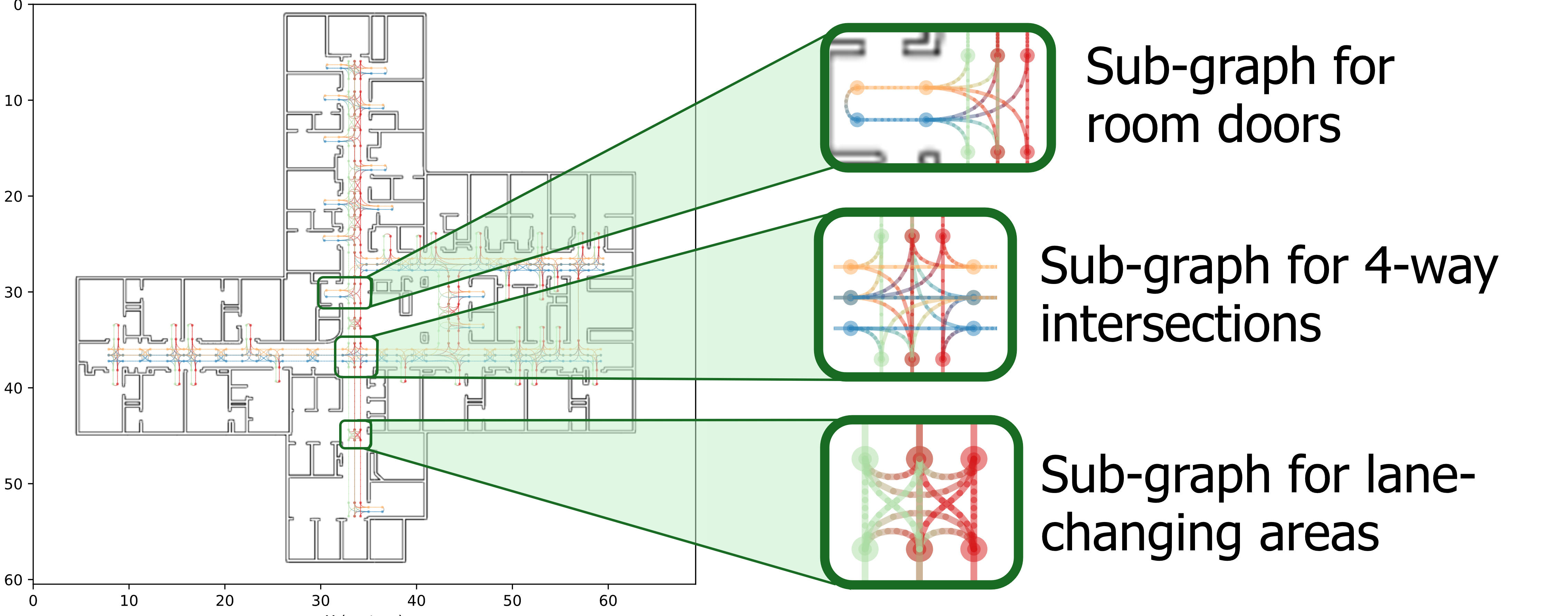} 
    \caption{Example traversal graph used in the case-study simulations. Waypoint nodes and feasible local motion edges are overlaid on the hospital navigation map, with zoomed regions showing graph connectivity in representative corridor and intersection areas. The resulting graph supports shortest-path travel-time estimation and graph-constrained routing for task assignment and scheduling.} 
    \label{fig:traversal_graph} 
\end{figure}

The simulated environment consists of inpatient hospital floors represented as traversal graphs. An example of how the hospital floor environments are generated is shown in \Cref{fig:env_generation}, while the extracted traversal graph for robot motion is shown in \Cref{fig:traversal_graph}. Nodes in the traversal graph correspond to traversable waypoints, patient rooms, supply rooms, and robot parking locations, while edges encode feasible robot motion through the floor. Each experiment uses the same graph representation and motion-planning interface for all policies. Shortest paths between any pair of nodes are cached for fast retrieval and fast heuristic travel-time estimation. Robots start from parking locations and must complete assigned service tasks while respecting routing, compatibility, timing, and return-to-station constraints.

We consider two robot families. Monitoring robots serve vital-sign requests, including blood pressure, heart rate, respiratory rate, temperature, and oxygen saturation tasks. Delivery robots serve medication requests, which require visiting a supply location before delivering to the patient-room endpoint. These two robot families define the heterogeneous team composition $\mathbf M=(M_z)_{\{z\in\mathcal Z\}}$ used in the problem formulation: monitoring robots and delivery robots provide different capabilities and cannot generally substitute for one another.

All policies are evaluated on four inpatient floors from the west wing of BIDMC. Each floor contains approximately $25$ to $30$ patient beds. The historical data span June 24, 2024 through June 29, 2025. For each floor, we select six ISO weeks for evaluation: two high-demand weeks, two medium-demand weeks, and two low-demand weeks. This produces $42$ test days per floor and $168$ floor-day evaluations per policy. Each floor-day is simulated over a $12$ hour window starting at $6am$ and ending at $6pm$, with one simulator time step corresponding to one second. The rest of the historical data is left as part of the training and validation sets. Request arrivals, service windows, patient-room endpoints, supply-room endpoints, and scheduled medication times are taken from the processed hospital historical data.

\subsubsection{Constructing the Test Set}
\label{subsubsec:case_study_test_set}

The held-out evaluation set is designed to test the policies under different demand regimes on each floor. We identify high-, medium-, and low-demand weeks using Laney $u$-charts \cite{montgomery2019statistical_quality_control}, a statistical process-control method for identifying unusual count rates when the number of observational units may vary over time. In this setting, the unit is a floor-day, so the weekly rate is the average number of requests per floor-day for a given floor and task type. The Laney $u$ chart adjusts the standard Poisson control limits using an estimated dispersion factor, making the demand-regime labels less sensitive to natural overdispersion or underdispersion in hospital request data \cite{laney2002_improved_attribute_charts}.

\begin{figure}
    \centering
    \includegraphics[width=\linewidth]{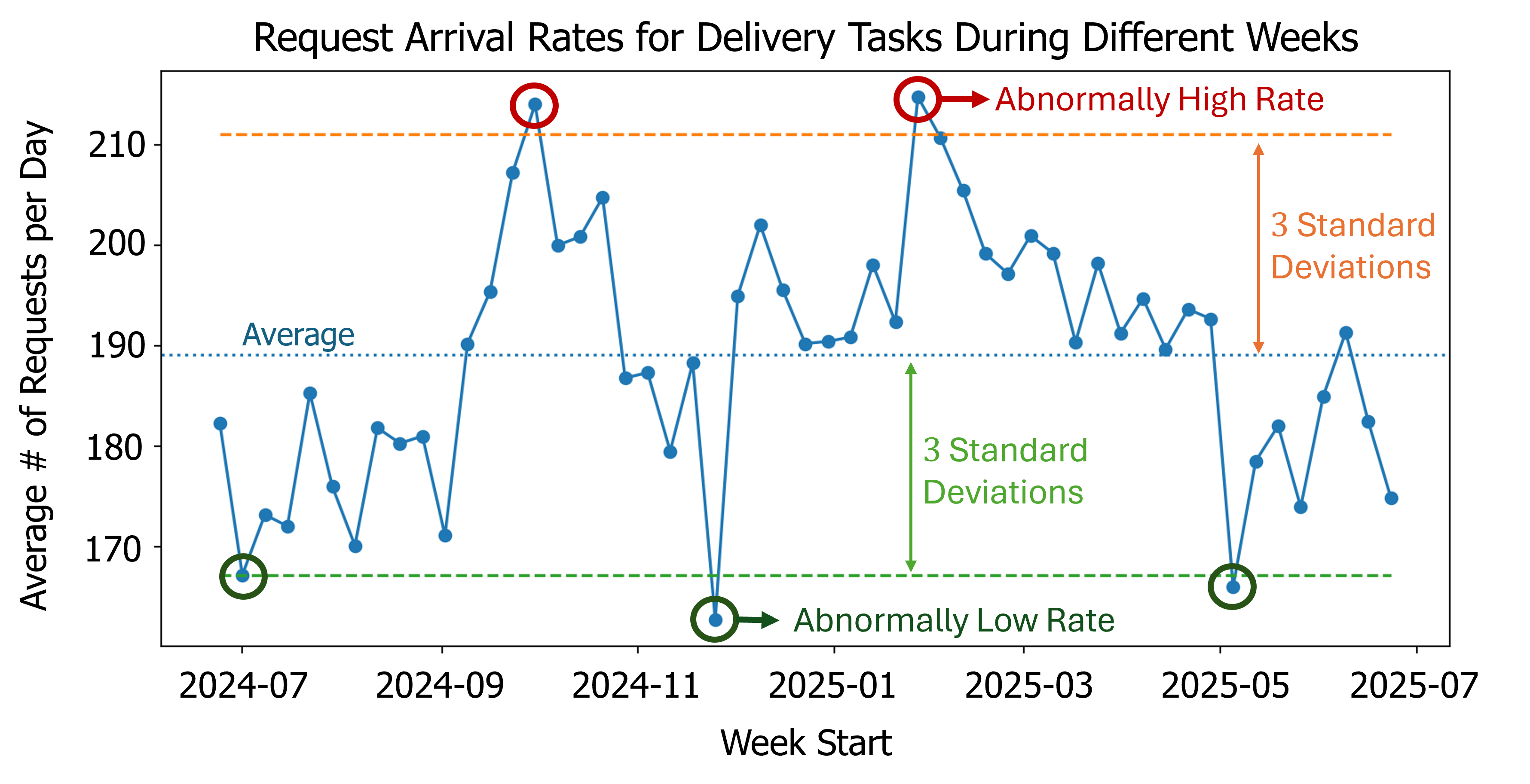}
    \caption{Example weekly delivery-request arrival rates used to select evaluation periods. The empirical mean is shown by the dotted blue line, and the dashed lines mark the Laney-adjusted upper and lower control limits. Weeks exceeding these limits are identified as abnormal high- or low-demand periods and are used to test the robustness of the proposed planner under demand shifts.}
    \label{fig:laney_u_chart}
\end{figure}

For each floor and task type, we construct a weekly Laney $u$ chart over the historical data. A week is flagged as high demand for a task type if its request rate exceeds the upper control limit, and as low demand if its request rate falls below the lower control limit. An example $u$-chart with flagged weeks for delivery requests is shown in \Cref{fig:laney_u_chart}. We then aggregate flags across task types at the floor level. High-demand and low-demand weeks are selected as those with the largest number of corresponding task-type flags, while medium-demand weeks provide reference operating conditions between these extremes. This selection rule favors weeks in which the demand regime reflects a broad change in floor activity rather than an isolated anomaly in a single task type.

The selected weeks are listed in \Cref{tab:case_study_dataset_summary}. Because the split is defined at the week level, every policy is evaluated on the same contiguous request sequences for each floor and demand regime. This preserves temporal correlations in the request stream and avoids evaluating policies on isolated requests removed from their operational context.

\begin{table*}
\centering
\caption{Summary of the hospital-floor test set. The table reports the selected high-, medium-, and low-demand weeks for each floor. The final row gives the total number of requests in each demand level aggregated across all task types, selected weeks, and floors.}
\label{tab:case_study_dataset_summary}
\small
\begin{tabular}{|c|p{0.22\textwidth}|p{0.22\textwidth}|p{0.22\textwidth} |}
\hline
Floor
& High-demand weeks
& Medium-demand weeks
& Low-demand weeks \\ \hline
2
& 2025-W06, 2025-W08
& 2024-W44, 2025-W14
& 2024-W27, 2024-W28 \\ \hline
3
& 2025-W05, 2025-W10
& 2024-W44, 2025-W14
& 2024-W46, 2025-W02 \\ \hline
7
& 2024-W39, 2024-W40
& 2024-W44, 2025-W14
& 2024-W46, 2025-W26 \\ \hline
9
& 2024-W43, 2025-W06
& 2024-W44, 2025-W14
& 2025-W19, 2025-W21 \\ \hline
\textbf{Total requests}
& \textbf{36,948}
& \textbf{30,824}
& \textbf{25,300} \\ \hline
\end{tabular}
\end{table*}

\subsection{Implementation Details}
\label{subsec:case_study_implementation_details}

This subsection specifies the implementation choices and parameter values used to instantiate the proposed framework in the hospital-floor case study. The goal is not to restate the algorithms in \Cref{sec:approach}, but to describe how their inputs, hyperparameters, heuristics, and motion-planning components are configured for this domain.

Both monitoring and delivery robots are modeled as disk robots with diameter $0.40,\mathrm{m}$. The two robot families have different capabilities: monitoring robots can service vital-sign requests, while delivery robots can service medication-delivery requests. Unless otherwise stated, both robot families use a nominal speed of $0.30,\mathrm{m/s}$. The traversal graph is constructed so that its edges correspond to physically realizable motions under the robot footprint and speed assumptions. All committed robot trajectories are planned using a variant of Safe Interval Path Planning with reservation tables, denoted SIPPwRT \cite{ma2019_sippwrt}, which extends SIPP \cite{phillips2011_sipp} time-interval reasoning with reservations induced by previously planned robot trajectories \cite{silver2005_cooperative_pathfinding}. This choice is well suited to prioritized multi-robot routing because each robot can be planned sequentially while treating earlier planned trajectories as dynamic reservations.

\subsubsection{One-Robot-at-a-Time Rollout Configuration}
The one-robot-at-a-time rollout rule in \Cref{subsec:oat_prediction_aware_rollout} requires an ordering of robots at each decision time. In the case study, the ordering set $\mathcal O_t$ is constructed from the current simulator state by sorting robots according to the time at which they are expected to become available. Robots that are already available and unassigned are considered first, followed by robots whose currently assigned request is expected to finish within two minutes. Robots with longer remaining service times are not included in $\mathcal O_t$ at that decision time. If $\mathcal O_t=\emptyset$, no new assignments are made and the current schedules continue to execute.

The rollout depth is set to one hour of simulated operation. Since the simulator advances in one-second time steps, this corresponds to $D = 3600$. Thus, each rollout evaluation accounts for predicted and scheduled requests that may enter during the next hour, while avoiding the computational cost of simulating to the end of the full operating window.

\subsubsection{Candidate-Control Generation}
The candidate generator in \Cref{subsec:candidate_control_generation} retains at most $L=20$ assignment actions per processed robot. The wait action is then added separately, so the robot-local candidate set contains at most $L+1$ controls. The value $L=20$ was chosen to preserve the most urgent and promising assignments while keeping the number of rollout evaluations manageable.

The candidate-ranking heuristics are computed as follows. The travel heuristic $F_{\ell,j}^{\mathrm{travel}}$ is obtained from cached shortest-path distances on the traversal graph. Specifically, the shortest-path distance needed to traverse the service-node sequence $\boldsymbol{\rho}_j$ is divided by the nominal speed of robot $\ell$. This computation ignores dynamic robot-robot interactions and service-node conflicts; those are handled by SIPPwRT for committed paths and by the reservation-table approximation in the rollout base policy.

The reach estimate $F_{\ell,j,t}^{\mathrm{reach}}$ used in the heuristic completion calculation is computed with the motion planner. Candidate actions that cannot satisfy the request deadline under this heuristic estimate are discarded before rollout evaluation.

\subsubsection{Future-Request Scenario Generation}
Future request scenarios are generated from the temporal point process models described in \Cref{subsec:future_request_generation}. In the hospital case study, we instantiate one local TPP model per patient process. At each rollout decision, the planner draws $S=20$ future-request samples. This value was chosen to represent variation in the predictive distribution while keeping online rollout runtimes reasonable.

Because each TPP prediction is associated with a patient process, the graph-grounded service nodes can be derived directly from the patient's room information at the sampled time. For vital-sign requests, the patient room is the service endpoint. For medication requests, the service-node sequence includes the relevant medication supply location followed by the patient-room endpoint.

We evaluate two mark representations for the temporal point process models. In the standard mark representation, vital-sign events are marked only by the request type, such as blood pressure, temperature, oxygen saturation, heart rate, or respiratory rate. In the enhanced mark representation, vital-sign events include both the request type and a discretized representation of the measured value, so that the model can condition on the magnitude of recent patient measurements.

Medication names are first mapped to RxNorm identifiers \cite{nlm_rxclass_atcprod_2026}. The medication mark is then constructed from Anatomical Therapeutic Chemical (ATC) codes \cite{who_atc_ddd_index_2026}. In the standard medication mark representation, we use the ATC level-3 code. In the enhanced medication mark representation, we use the first three ATC hierarchy components, which provide a richer representation of the medication class while avoiding excessive sparsity at the individual-drug level.

Predicted requests that match scheduled or already observed requests are removed before rollout evaluation. The matching tolerance is set to $\Delta = 10$ minutes. Thus, a sampled prediction is removed if it corresponds to a request of the same task context that is already known to the simulator within a (10)-minute time window. This prevents the rollout estimator from double-counting demand that is already represented in the state.

\subsubsection{Interaction-Aware Base Policy}
The base policy used inside rollout is the interaction-aware future-cost estimator described in \Cref{subsec:greedy_base_policy}. Its reservation table is a simplified version of the full SIPPwRT reservation table used for committed motion planning. Instead of reserving all vertices and edges along complete robot trajectories, the base-policy reservation table only records service intervals at patient rooms and medication supply rooms. This approximation captures the dominant service-location interactions while avoiding the cost of planning full collision-free paths for every hypothetical future request in every rollout sample.

For each simulated assignment considered by the base policy, estimated service times are computed using cached graph shortest paths, the nominal robot speed, request execution durations, release times, deadlines, and the simplified service-node reservation table. If a tentative service interval overlaps a reservation at the same patient room or supply room, the interval is shifted to the earliest available time that satisfies the request's service-window constraints. If no feasible interval exists before the request deadline, the request is marked as rejected or no longer serviceable in the simulated state.

\subsubsection{Adaptive Prediction Confidence}
The adaptive prediction-confidence mechanism in \Cref{subsec:adaptive_prediction_confidence} is enabled for rollout variants with prediction reweighting. Prediction weights are updated in bins of length $\Delta_\lambda = 5\text{ minutes}$. The minimum prediction weight is $\lambda_{\min}=0.05$, so predicted requests can be strongly downweighed but are not completely ignored. The exponential smoothing parameters are $\alpha_{\mathrm{over}}=0.3$ and $\alpha_{\mathrm{time}}=0.2$. Thus, the overprediction state reacts somewhat faster than the timing-error state. The confidence-decay parameters are $\beta_{\mathrm{over}}=4.0$ and $\beta_{\mathrm{time}}=1.0$, which makes persistent overprediction reduce confidence more strongly than one-bin timing shifts. The numerical constant used for normalization and logarithmic clipping is $\epsilon=10^{-6}$. Hierarchical fallback is enabled in all rollout variants that use adaptive prediction confidence. The local-context credibility prior is $\zeta=3.0$. Therefore, patient-specific confidence weights are initially blended strongly with the broader fallback context, and they dominate only after enough positive-prediction windows have been observed. Confidence updates are skipped for zero-prediction windows, so contexts are not penalized or rewarded when the prediction snapshot contains no positive predicted mass.

\subsubsection{Scheduled Requests, Re-optimization, and Decision Triggers}
Known future scheduled requests are included in the rollout future-cost estimate for the proposed rollout variants. They are treated as known commitments and receive unit weight in the weighted schedule score. However, they are not added to the immediately assignable request set until they enter the simulator state. In contrast, stochastic TPP predictions are never eligible as immediate assignment actions; they affect decisions only through sampled rollout simulations and confidence-weighted future costs.

Selective re-optimization is enabled only in rollout variants that include the mechanism described in \Cref{subsec:reoptimization}. When enabled, assigned but unstarted requests may be released back to the pending set if newly observed requests have smaller slack. Started requests are never released or reassigned.

\subsection{Temporal-Point-Process Prediction Ablation}
\label{subsec:tpp_ablation}

Before evaluating the online assignment policies, we evaluate the temporal point process models used to generate future request scenarios in \Cref{subsec:future_request_generation}. This ablation serves two purposes. First, it identifies the prediction model configuration used by the rollout policies in the remaining experiments. Second, it tests whether richer timed-mark representations and previous-day context improve the quality of sampled future request sequences in the hospital setting.

The ablation compares the candidate temporal point process models used to sample future request sequences. The model set includes: RMTPP \cite{Du2016}, NHP \cite{mei2017neuralhawkes}, FullyNN \cite{Omi2019}, SAHP \cite{Zhang2020}, THP \cite{Zuo2020}, IntesityFree \cite{Shchur2020Intensity-Free}, AttnNHP \cite{mei2022transformer}, ANHN \cite{Xue2021}, WSM-THP \cite{cao2024is}, S2P2 \cite{chang2026deep}, TriTPP \cite{shchur2020fast}, inhomogeneous Poisson \cite{shchur2020fast}, Renewal \cite{shchur2020fast} , Modulated Renewal \cite{shchur2020fast}, Spline Transformer \cite{shchur2020fast}, and FlexTPP-based variants \cite{draxler2026transformers}. The implementation of the first 8 methods was derived from \cite{xue2024easytpp}, while the implementation for the rest of the methods was derived from their respective papers. The only modifications applied to all architectures were dimensionality adjustments needed to process the enhanced marks. All models under consideration were tested for both enahnced and standard marks. The standard mark representation uses only the request type for vital-sign requests and a coarser medication-code representation for medication requests. The enhanced mark representation augments vital-sign marks with discretized measurement information and uses a richer medication-code representation, as described in \Cref{subsec:case_study_implementation_details}.

For the FlexTPP variants, we also compare models trained with and without previous-day summary conditioning. Previous-day conditioning summarizes the recent patient-level history available before the prediction prefix. For vital-sign requests, this includes whether previous-day context is available, the time since the last previous-day request, total previous-day request count, task-specific request counts, task-specific last request times, and summary statistics of previous-day measurements. For medication requests, the conditioning vector summarizes previous-day vital-sign and medication-administration activity for the same local patient process.

We train all architectures for $300$ epochs using the training set. We select the log likelihood as the optimization objective for each temporal point process, and we use Adam \cite{Kingma2014AdamAM} as the optimizer. The learning rate for each architecture is chosen using a hyper-parameter search. We choose a batch size of $32$, and we shuffle the batches at every epoch during training. The number of layers and the width of each layer for each architecture are chosen following the recommendations given in their respective papers. All training was done using two NVIDIA RTX A6000 ADA. 

Prediction quality is evaluated by comparing sampled future event sequences against the realized future sequence. For a local process $p$ and prediction prefix ending at time $t$, let
\begin{equation}
\widehat W_{p,t}^{s}
=
\left(
\widehat e_{1}^{s},
\ldots,
\widehat e_{\widehat N_s}^{s}
\right),
\qquad
W_{p,t}
=
\left(
e_1,
\ldots,
e_N
\right)
\label{eq:otd_predicted_and_true_sequences}
\end{equation}
denote the $s$-th sampled predicted suffix and the corresponding realized suffix, where $\widehat N_s$ is the number of predicted events and $N$ is the actual number of observed events. Each event is a timed mark $e=(\tau,k,\kappa)$, where $\tau$ is the event time, $k$ is the request type, and $\kappa$ is the model-specific mark representation. We evaluate each sampled suffix using a marked Optimal Transport Distance (OTD), implemented as an ordered edit-alignment distance between $\widehat W_{p,t}^{s}$ and $W_{p,t}$.

Matching a predicted event $\widehat e_i^s$ to a realized event $e_j$ incurs the substitution cost
\begin{equation}
d_{\mathrm{sub}}
\left(
\widehat e_i^s,
e_j
\right)
=
\alpha_{\mathrm{OTD}}
\frac{
\left|
\widehat \tau_i^s-\tau_j
\right|
}{
\tau_{k_j}^{\mathrm{scale}}
}
+
\beta_{\mathrm{OTD}}
\mathbb 1_{
\left\{
\widehat k_i^s\neq k_j
\right\}},
\label{eq:otd_substitution_cost}
\end{equation}
where $\tau_{k_j}^{\mathrm{scale}}$ is a task-specific time scale used to normalize timing errors. Predicted events that cannot be matched to realized events incur deletion cost $c_{\mathrm{del}}$, and realized events that are missing from the prediction incur insertion cost $c_{\mathrm{ins}}$.

The sequence-level OTD is
\begin{align}
\operatorname{OTD}
\left(
\widehat W_{p,t}^{s},
W_{p,t}
\right)
&=
\min_{\mathcal A}
\Biggl[
\sum_{(i,j)\in\mathcal A_{\mathrm{match}}}
d_{\mathrm{sub}}
\left(
\widehat e_i^s,
e_j
\right)
\nonumber \\
&\quad
+
c_{\mathrm{del}}
|\mathcal A_{\mathrm{del}}|
+
c_{\mathrm{ins}}
|\mathcal A_{\mathrm{ins}}|
\Biggr],
\label{eq:otd_sequence_metric}
\end{align}
where $\mathcal A$ ranges over all ordered edit alignments between the sampled and realized suffixes. Lower OTD values indicate better agreement in event timing, task type, and sequence length. For each prediction prefix, we estimate the expected OTD over sampled futures by
\begin{equation}
\widehat{\operatorname{OTD}}_{p,t}
=
\frac{1}{S_{\mathrm{eval}}}
\sum_{s=1}^{S_{\mathrm{eval}}}
\operatorname{OTD}
\left(
\widehat W_{p,t}^{s},
W_{p,t}
\right),
\label{eq:expected_otd_estimator}
\end{equation}
where $S_{\mathrm{eval}}$ is the number of sampled futures used for the prediction evaluation. 

For all OTD results reported below, we use $S_{\mathrm{eval}} = 20$, $\alpha_{\mathrm{OTD}}=1.0$, $c_{\mathrm{del}}=1.0$, and $c_{\mathrm{ins}}=1.0$. The timing weight $\alpha_{\mathrm{OTD}}$ controls the contribution of normalized timing error for matched events. The deletion penalty $c_{\mathrm{del}}$ penalizes spurious predicted events, and the insertion penalty $c_{\mathrm{ins}}$ penalizes realized events that were missed by the prediction. The normalizing constant $\tau_{k_j}^{\mathrm{scale}}$ is set to the mean inter-event time for request type $k_j$ in the training data.

For monitoring requests, we use a finite request-type substitution penalty $\beta_{\mathrm{OTD}}=0.25$. This allows the alignment to match different vital-sign request types at a small cost, reflecting that the same monitoring robot can service all vital-sign tasks. The resulting OTD decomposes into timing, request-type mismatch, deletion, and insertion components. For medication-delivery requests, we use hard type matching: substitutions between different medications are disallowed. This reflects that different medication requests should not be treated as interchangeable. Therefore, the delivery OTD decomposes into timing, deletion, and insertion components, without a finite request-type mismatch component.

\begin{figure*}
\centering
\includegraphics[width=0.8\linewidth]{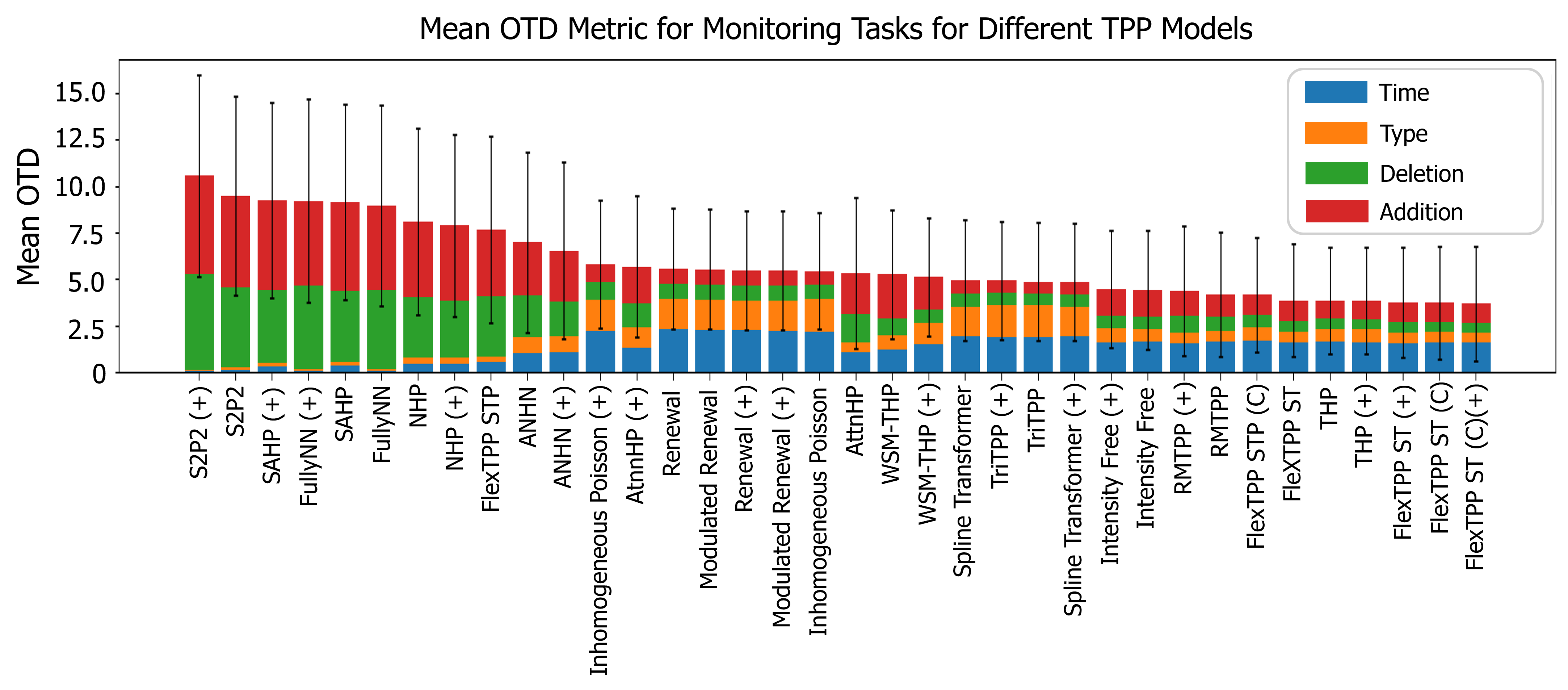}
\caption{\small{Comparison of temporal point process (TPP) models for monitoring-request prediction using the OTD metric. Each stacked bar reports mean OTD for one model configuration, decomposed into timing, request-type, deletion, and insertion components. Error bars indicate variability across evaluation samples. Lower values indicate better agreement between predicted and observed request sequences, while smaller deletion and insertion components correspond to fewer missed or spurious predicted events. TPP model names and references are given at the beginning of \Cref{subsec:tpp_ablation}. In the model labels, $(+)$ denotes the use of enhanced marks, $(C)$ denotes conditioning on previous-day information, and $(C)(+)$ denotes the use of both enhanced mark and previous day conditioning.}}
\label{fig:otd_monitoring}
\end{figure*}

\begin{figure*}
\centering
\includegraphics[width=0.8\linewidth]{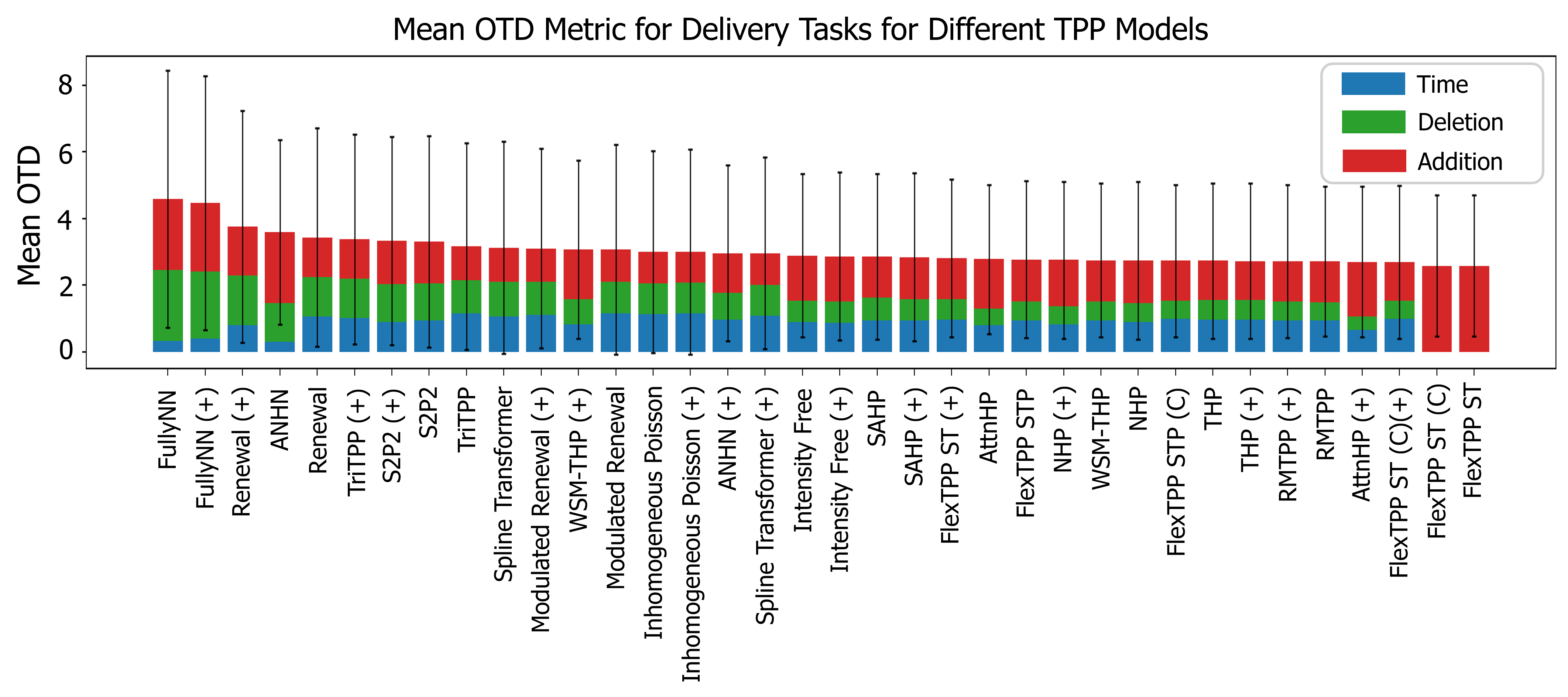}
\caption{\small{Comparison of temporal point process (TPP) models for medication-delivery prediction using the OTD metric. Each stacked bar reports mean OTD for one model configuration, decomposed into timing, request-type, deletion, and insertion components. Error bars indicate variability across evaluation samples. Lower values indicate better agreement between predicted and observed request sequences, while smaller deletion and insertion components correspond to fewer missed or spurious predicted events.TPP model names and references are given at the beginning of \Cref{subsec:tpp_ablation}. In the model labels, $(+)$ denotes the use of enhanced marks, $(C)$ denotes conditioning on previous-day information, and $(C)(+)$ denotes the use of both enhanced mark and previous day conditioning.}}
\label{fig:otd_delivery}
\end{figure*}

The OTD evaluation is stratified by the same high-, medium-, and low-demand weeks used in the assignment-policy experiments. Because the qualitative conclusions were similar across load regimes, \Cref{fig:otd_monitoring,fig:otd_delivery} report the high-demand results, where prediction errors are most likely to affect downstream assignment decisions. \Cref{fig:otd_monitoring} summarizes monitoring-request prediction, and \Cref{fig:otd_delivery} summarizes medication-delivery prediction.

For monitoring requests, \Cref{fig:otd_monitoring} shows that the FlexTPP model with enhanced marks and previous-day context obtains the lowest OTD among the evaluated configurations. This result indicates that measurement-aware marks and recent patient-level history provide useful signal for predicting future vital-sign request sequences, provided that the model architecture can use the additional conditioning information effectively.

For medication-delivery requests, \Cref{fig:otd_delivery} shows that FlexTPP-based models also outperform the other evaluated prediction models. However, unlike the monitoring case, the enhanced mark representation and previous-day context do not consistently improve OTD. One likely explanation is that medication-code features are substantially sparser than vital-sign marks, so the richer representation increases feature granularity without providing enough repeated evidence for the model to exploit. Based on these results, the rollout experiments use the best-performing monitoring and medication prediction configurations identified by this ablation to generate future request scenarios.

\subsection{Team-Sizing Results}
\label{subsec:case_study_team_sizing_results}

We next evaluate the historical team-composition procedure described in \Cref{subsec:team_sizing}. This step determines the fixed heterogeneous team $\mathbf M$ used in the subsequent online policy experiments. The procedure is important because monitoring and delivery robots provide different capabilities: additional delivery robots cannot resolve monitoring bottlenecks, and additional monitoring robots cannot resolve medication-delivery bottlenecks. The sizing procedure must therefore select enough capacity for each robot family while also verifying that the resulting joint team composition is feasible when all request types, timing constraints, and routing interactions are evaluated together.

\begin{figure}
\centering
\includegraphics[width=\linewidth]{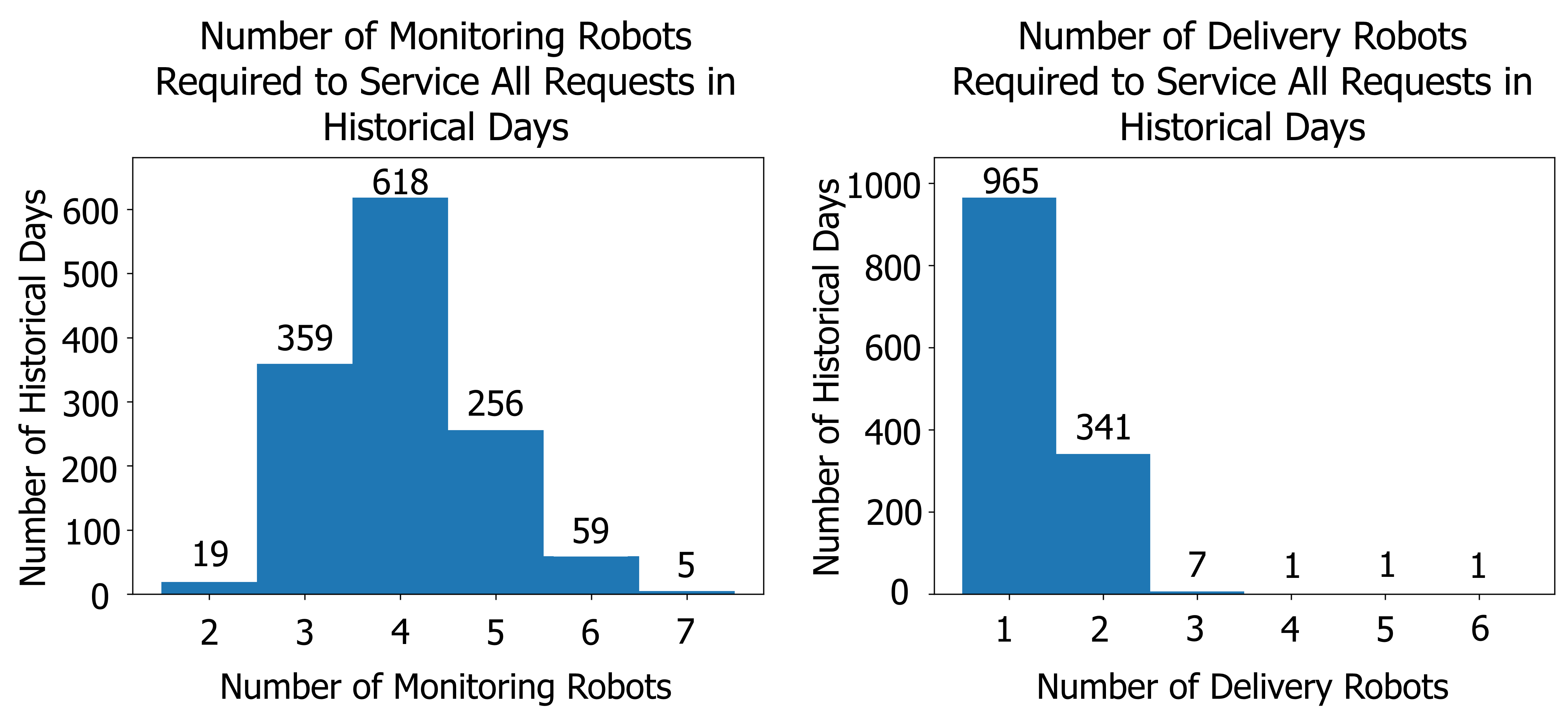}
\caption{\small{Day-wise robot requirements obtained during team-size initialization. For each historical day, the sizing routine increases the number of monitoring or delivery robots until all requests are feasibly served. The histograms summarize the resulting feasible robot counts across historical days and are used to initialize the heterogeneous team composition at the desired empirical percentile.}}
\label{fig:team_sizing}
\end{figure}

\Cref{fig:team_sizing} shows the day-wise robot requirements produced by the histogram-based initialization phase in \Cref{algo:team_sizing_initialization}. For each historical floor-day, the sizing routine increases the number of robots in the relevant family until all requests of that family can be served feasibly. The monitoring histogram summarizes the number of monitoring robots required across historical floor-days, and the delivery histogram summarizes the corresponding requirement for medication-delivery robots. Using the $99$-th percentile of these empirical distributions gives an initial team composition of $6$ monitoring robots and $2$ delivery robots.

The initialized composition is then checked using the joint historical verification phase described in \Cref{algo:team_sizing}. \Cref{tab:team_sizing_summary} summarizes the composition before and after this verification step. The monitoring fleet remains unchanged at $6$ robots, while the delivery fleet increases from $2$ to $3$ robots.

\begin{table}
\centering
\caption{Selected heterogeneous team composition.}
\label{tab:team_sizing_summary}
\small
\begin{tabularx}{\linewidth}{|>{\raggedright\arraybackslash}X|c|c|}
\hline
Phase & Monitoring robots & Delivery robots \\ \hline
Histogram initialization & 6 & 2 \\ \hline
Joint historical verification & 6 & 3 \\ \hline
\end{tabularx}
\end{table}

The increase in delivery capacity illustrates why the verification phase is necessary: marginal percentile estimates for each robot family do not necessarily guarantee that the combined team achieves the target empirical feasibility level once shared timing constraints, routing interactions, and multi-request schedules are evaluated jointly. The verified team composition $\mathbf M=(6,3)$ is used as the fixed heterogeneous robot team in the remaining policy-comparison and ablation experiments, so subsequent performance differences reflect the online assignment policies rather than differences in available fleet capacity.

\subsection{Policy Comparison}
\label{subsec:case_study_comparative_results}

We next compare the proposed adaptive rollout policy against reactive, deadline-aware, prediction-based, and myopic assignment baselines. All policies are evaluated on the same held-out floor-days defined in \Cref{subsubsec:case_study_test_set}, using the fixed heterogeneous team composition selected in \Cref{subsec:case_study_team_sizing_results}. Thus, differences in performance reflect the online assignment policy rather than differences in fleet capacity, request streams, traversal graphs, or robot capabilities.

The main proposed method is the adaptive rollout policy with future scheduled requests, temporal-point-process prediction scenarios, adaptive prediction reweighting, and selective re-optimization enabled. This policy uses predicted requests only inside rollout simulations; immediate assignment actions are restricted to requests that have already entered the system.

\subsubsection{Baselines for Comparison}
\label{subsubsec:case_study_baselines}

The comparison includes the following policies.

\paragraph{Fleet Manager.}
The Fleet Manager baseline is a reactive queue-based policy adapted from \cite{Popolizio2024Fleet}. Requests are processed according to their release times. When a compatible robot is available, the policy assigns the next request to the available robot that can reach it most quickly. A collision-free path is then planned for the assignment, while previously committed paths are treated as fixed and immutable. This baseline represents a simple operational dispatching strategy that does not explicitly reason about future requests.

\paragraph{Token Passing (TP).}
The Token Passing baseline follows the multi-agent task-allocation framework of \cite{Ma2017TP}. Robots request a shared token when they are available. When a robot holds the token, it selects an unassigned task using a local cost based on travel distance to the task, without explicitly accounting for deadline urgency or future demand. After the task is selected, the robot plans a path to the task location while avoiding the trajectories of other robots. This baseline captures a standard decentralized assignment strategy with prioritized path planning.

\paragraph{Token Passing with Task Swaps (TPTS).}
The Token Passing with Task Swaps baseline extends token passing by allowing unstarted tasks to be reassigned when a different robot can serve them more effectively \cite{Ma2017TP}. When a robot considers a task that is already assigned but not yet started, the policy may remove that assignment and assign the task to the current robot, provided that the displaced robot can obtain another feasible assignment. This baseline introduces limited reassignment flexibility while remaining reactive and myopic.

\paragraph{Token Passing with Deadlines (TP-D).}
The deadline-aware token-passing baseline incorporates request deadlines into the task-selection objective \cite{Makino2024Deadlines}. Instead of ranking tasks only by travel cost, the policy uses a weighted score that combines travel cost and deadline urgency. We set the deadline-weight parameter to $\alpha=0.2$, which was selected by grid search over $\alpha\in{0.1,0.2,0.3,0.4}$ on the tuning runs. This baseline tests whether explicit deadline awareness improves service reliability relative to distance-based token passing.

\paragraph{Deadline-Aware Token Passing with Task Swaps (D-TPTS).}
This baseline combines deadline-aware task scoring with reassignment of unstarted tasks \cite{Makino2024Deadlines}. Newly available robots request the token, evaluate deadline-aware assignment costs, and may deallocate lower-priority unstarted tasks when a more urgent task arrives. We again use $\alpha=0.2$, selected from the same grid search. This policy is the strongest token-passing baseline because it combines deadline awareness with limited schedule repair.

\paragraph{Idle Rebalance.}
The Idle Rebalance baseline uses predictions of future requests to reposition idle robots \cite{Fan2023IdlePred}. When robots are idle, predicted requests can act as pseudo-tasks that influence robot positioning. This differs from the proposed rollout policy: in our method, predicted requests are never immediate assignment actions and affect decisions only through the simulated future-cost estimate.

\paragraph{Greedy Assignment Policy (Base Policy).}
The greedy assignment baseline is the standalone myopic policy that we proposed as the base policy for rollout in \Cref{subsec:greedy_base_policy}. It orders currently known requests by urgency, evaluates compatible robots, and assigns the request to the robot that results in the lowest heuristic assignment cost.

\paragraph{Proposed Adaptive Rollout (Our Approach).}
The proposed method is the prediction-aware adaptive rollout policy developed in \Cref{sec:approach}. It restricts immediate candidate actions to currently observed eligible requests, evaluates each candidate through sampled future request scenarios, scores predicted requests using adaptive confidence weights, and selectively re-optimizes assigned but unstarted requests when newly observed demand makes the current schedule undesirable.

\subsubsection{Performance Metrics}
\label{subsubsec:case_study_performance_metrics}

We evaluate policies using service-level, service-quality, and computational metrics.

The service-level metrics are the number of serviced and rejected requests. A request is counted as serviced if it is assigned to a compatible robot and completed within its service window. A request is counted as rejected if the policy cannot assign it feasibly before its deadline. We report these counts separately for high-, medium-, and low-demand days, because policy differences are expected to be most pronounced when robot capacity and request deadlines are most constrained.

The service-quality metrics are computed over requests that are successfully serviced. For each serviced request $r_j$, the wait time is $h_j = \max\{0,C_j-t_j^{\mathrm{des}}\}$, 
where $C_j$ is the realized completion time and $t_j^{\mathrm{des}}$ is the desired service time. For each policy and day, we compute the mean serviced-request wait time and the $95$-th percentile serviced-request wait time. We then plot the distribution of these daily statistics across days, stratified by demand regime. The mean wait-time plots summarize typical service responsiveness, while the $95$-th percentile plots capture tail delays.

The computational metric is the planning time required to generate a decision for one request. We report the distribution of per-request planning times across the evaluation runs. This metric is important because the proposed rollout policy evaluates multiple candidate actions over sampled future scenarios, whereas the baselines generally make more local decisions.

\subsubsection{Service-Level Results}
\label{subsubsec:case_study_service_level_results}

\Cref{tab:service_level_results} reports the number of serviced and rejected requests for each policy, stratified by demand regime. This metric evaluates whether each policy can maintain feasibility with the heterogeneous team selected in \Cref{subsec:case_study_team_sizing_results}. High-demand days are the most informative for service-level performance because robot availability and deadline constraints are most likely to become binding. Medium- and low-demand days test whether the same policies remain feasible when the system is less capacity constrained.

\begin{table*}
\centering
\caption{Service-level results by demand regime. Each entry reports serviced and rejected requests for the corresponding policy and load level.}
\label{tab:service_level_results}
\small
\begin{tabular}{|p{0.27\textwidth}|cc|cc|cc|}
\hline
\multirow{2}{*}{Policy}
& \multicolumn{2}{c|}{High demand}
& \multicolumn{2}{c|}{Medium demand}
& \multicolumn{2}{c|}{Low demand} \\ 
\cline{2-7}
& Serviced & Rejected
& Serviced & Rejected
& Serviced & Rejected \\ 
\hline
Fleet Manager & 36,948 & 0 & 30,824 & 0 & 25,300 & 0 \\ \hline
Token Passing & 36,920 & 28 & 30,824 & 0 & 25,300 & 0 \\ \hline
Token Passing with Task Swaps & 36,920 & 28 & 30,824 & 0 & 25,300 & 0 \\ \hline
Token Passing with Deadlines & 36,948 & 0 & 30,824 & 0 & 25,300 & 0 \\ \hline
Deadline-Aware Token Passing with Task Swaps & 36,948 & 0 & 30,824 & 0 & 25,300 & 0 \\ \hline
Idle Prediction & 36,948 & 0 & 30,824 & 0 & 25,300 & 0 \\ \hline
Greedy Assignment Policy & 36,948 & 0 & 30,824 & 0 & 25,300 & 0 \\ \hline
Proposed Adaptive Rollout & 36,948 & 0 & 30,824 & 0 & 25,300 & 0\\ \hline
\end{tabular}
\end{table*}

The selected team composition is sufficient to service all requests for most policies across all demand regimes. The only service-level failures occur on high-demand days for Token Passing and Token Passing with Task Swaps, which reject $28$ requests out of $36{,}948$ high-demand requests. These rejections do not occur for the deadline-aware token-passing variants, the Fleet Manager baseline, Idle Prediction, the Greedy Assignment Policy, or the proposed adaptive rollout policy.

This result has two implications. First, it supports the fleet-sizing procedure: the selected team composition provides enough monitoring and delivery capacity to avoid persistent overload on the held-out floor-days. Second, it shows that aggregate rejection counts alone do not fully distinguish the stronger policies in this experiment, because most policies achieve zero rejections once the fleet is properly sized. The remaining comparisons therefore focus on service quality and computational cost. In particular, the wait-time results below evaluate whether policies that service the same number of requests differ in how close they complete those requests to their desired service times.

\subsubsection{Request Wait-Time Results}
\label{subsubsec:case_study_wait_time_results}

The service-level results in \Cref{tab:service_level_results} show that most policies complete nearly all requests under the selected team composition. We therefore use wait-time metrics to compare the quality of the schedules produced by the policies. For each serviced request $r_j$, wait time is computed as $h_j=\max{0,C_j-t_j^{\mathrm{des}}}$, where $C_j$ is the realized completion time and $t_j^{\mathrm{des}}$ is the desired service time. For each floor-day, we compute two daily summary statistics: the mean wait time over serviced requests and the $95$-th percentile wait time over serviced requests. The box plots below summarize the distribution of these daily statistics across evaluation instances for each demand regime.

\begin{figure*}
\centering
\includegraphics[width=\linewidth]{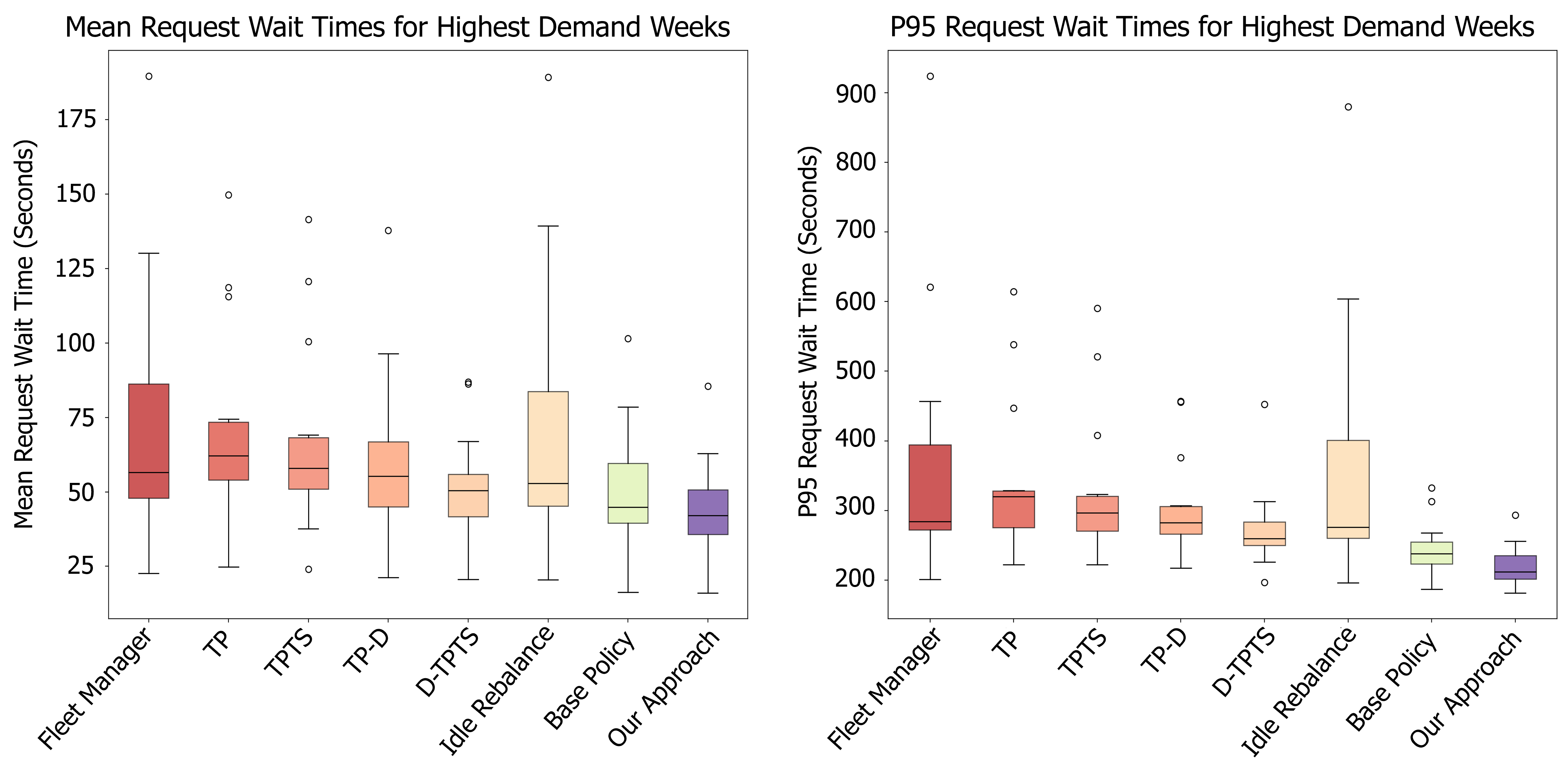}
\caption{\small{Comparison of request wait times across assignment policies during the highest-demand evaluation weeks. The left panel reports the distribution of daily mean wait times for serviced requests, and the right panel reports the distribution of daily $95$-th percentile wait times for serviced requests. Each box summarizes performance across evaluation instances, with lower values indicating shorter delays. Abbreviated policy labels correspond to the assignment policies defined in \Cref{subsubsec:case_study_baselines}.}}
\label{fig:comparison_results_high}
\end{figure*}

\Cref{fig:comparison_results_high} shows the wait-time comparison for the highest-demand weeks. In this regime, robot capacity and deadline constraints are most restrictive, so assignment decisions have the largest effect on downstream congestion. The proposed adaptive rollout policy achieves the lowest median wait time among the evaluated methods for both the daily mean and daily $95$-th percentile metrics. The median improvement is approximately $5\%$ for the mean wait-time metric and approximately $10\%$ for the $95$-th percentile metric. The reduction is even more pronounced in the upper tail: the proposed method reduces the $75$-th percentile and maximum values of both wait-time metrics by more than $15\%$ relative to the baselines.

These high-demand results indicate that the main benefit of prediction-aware rollout is not only a reduction in typical delay, but also a reduction in severe delay. This is consistent with the role of rollout: candidate assignments are evaluated based on their downstream effect on future capacity, which helps avoid decisions that are locally feasible but create later bottlenecks. The comparison with Idle Prediction is also informative. Although Idle Prediction uses future-demand information, it treats predicted requests as positioning targets. Its weaker performance shows that inaccurate or mistimed predictions can bias robot movement when predictions are treated too directly. The proposed rollout policy instead uses predictions as uncertain inputs to future-cost estimation, with adaptive weighting and re-optimization limiting the effect of unreliable forecasts.

\begin{figure*}
\centering
\includegraphics[width=\linewidth]{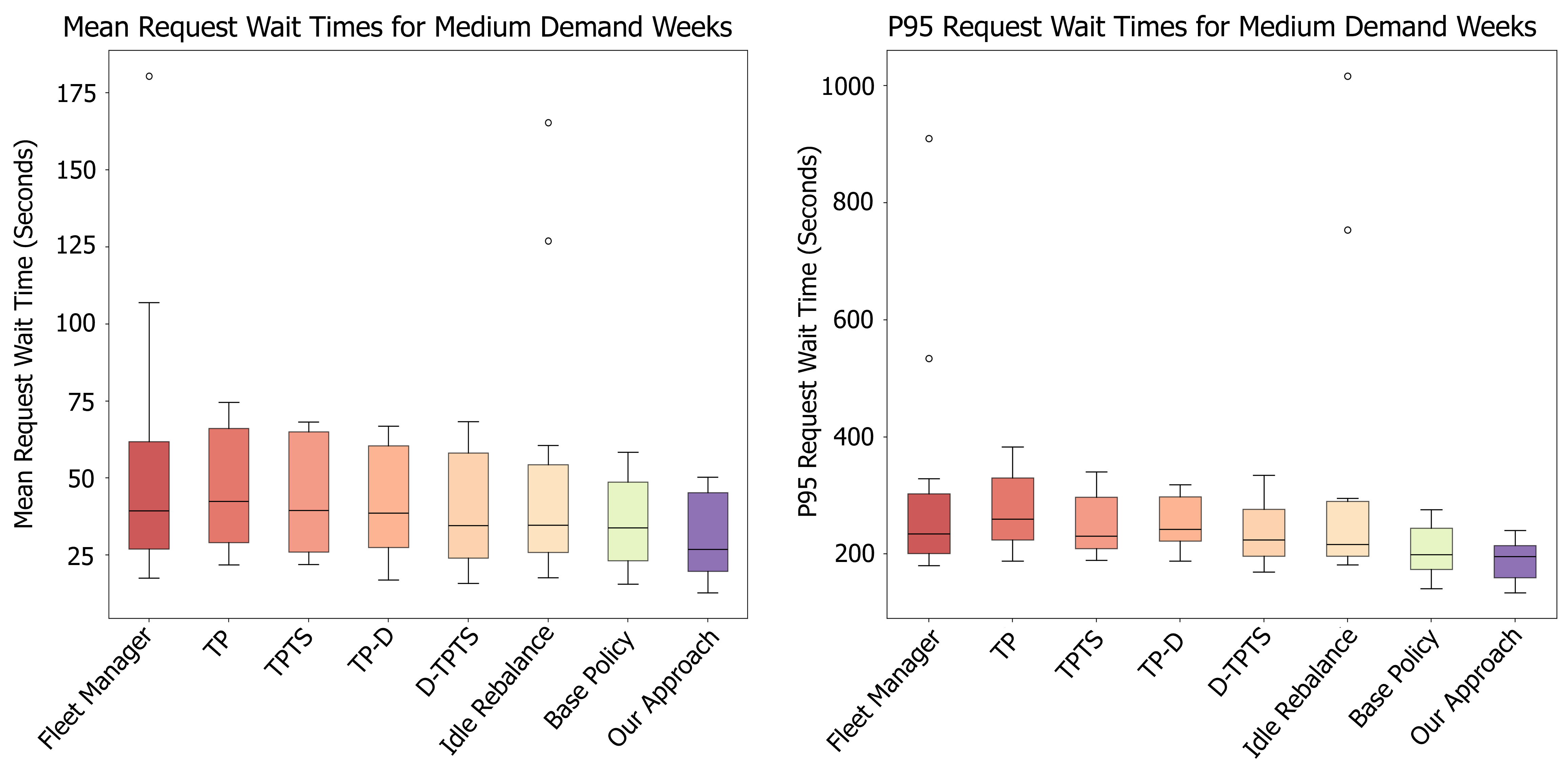}
\caption{\small{Comparison of request wait times across assignment policies during medium-demand evaluation weeks. The left panel reports the distribution of daily mean wait times for serviced requests, and the right panel reports the distribution of daily $95$-th percentile wait times for serviced requests. Each box summarizes performance across evaluation instances, with lower values indicating shorter delays. Abbreviated policy labels correspond to the assignment policies defined in \Cref{subsubsec:case_study_baselines}.}}
\label{fig:comparison_results_medium}
\end{figure*}

\Cref{fig:comparison_results_medium} reports the same comparison for medium-demand weeks. This regime represents nominal operating conditions, where the system is less congested than in the high-demand case but still has enough request volume for assignment quality to matter. The proposed adaptive rollout policy again obtains the lowest median wait time for both metrics. The median reduction is approximately $2\%$ for daily mean wait time and approximately $5\%$ for daily $95$-th percentile wait time. At the $75$-th percentile of the daily distributions, the improvement is approximately $5\%$ for the mean metric and approximately $15\%$ for the $95$-th percentile metric.

The medium-demand results show that rollout remains useful even when the system is not persistently overloaded. Under these conditions, future request predictions are generally informative enough to improve assignment timing, but direct prediction-based repositioning is still less effective than evaluating sampled futures through a scheduling-aware cost estimate. The proposed method benefits from this distinction: it uses multiple sampled futures and the interaction-aware base policy to estimate downstream cost, rather than greedily moving robots toward individual predicted requests.

\begin{figure*}
\centering
\includegraphics[width=\linewidth]{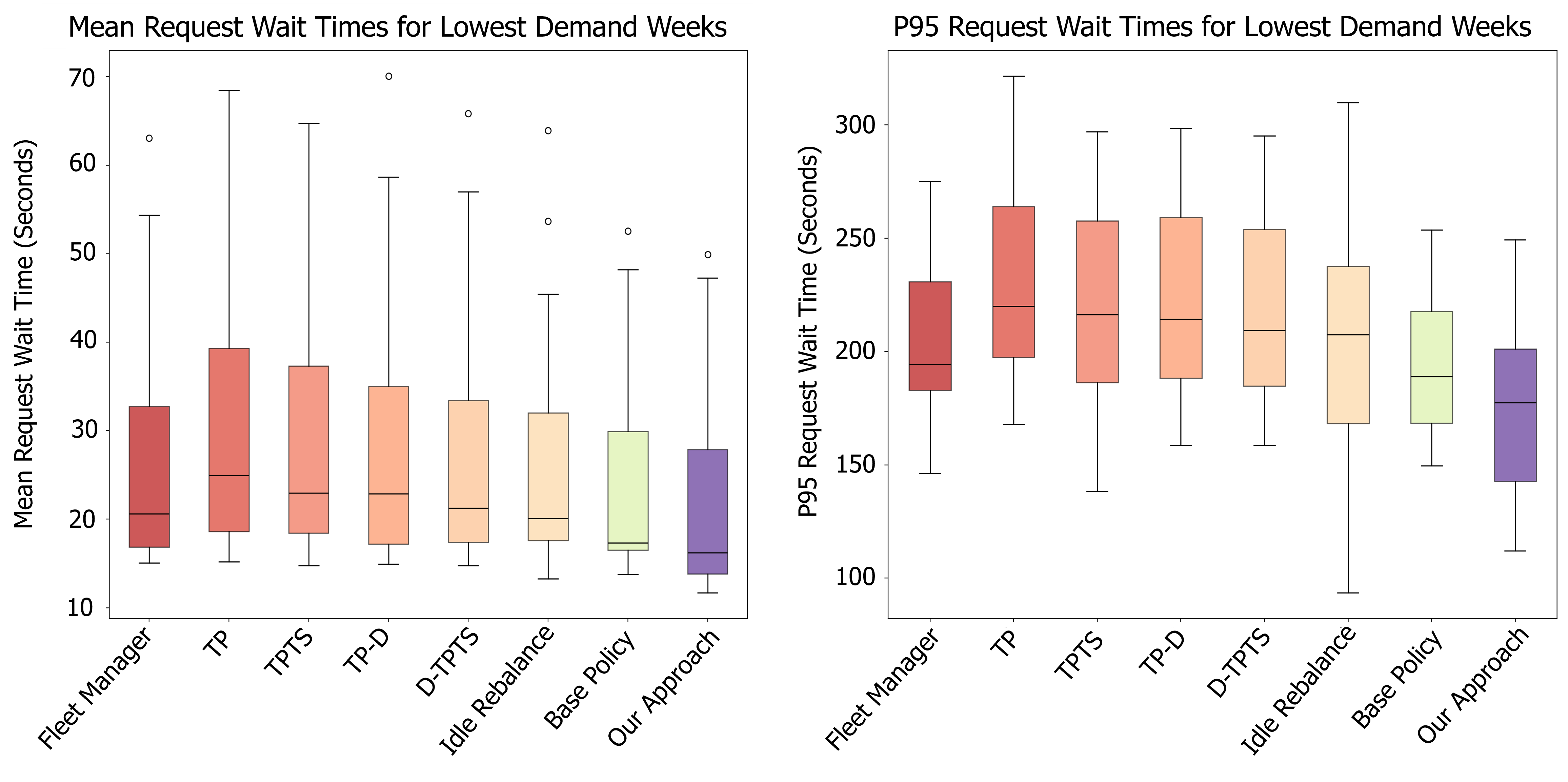}
\caption{\small{Comparison of request wait times across assignment policies during the lowest-demand evaluation weeks. The left panel reports the distribution of daily mean wait times for serviced requests, and the right panel reports the distribution of daily (95)-th percentile wait times for serviced requests. Each box summarizes performance across evaluation instances, with lower values indicating shorter delays. Abbreviated policy labels correspond to the assignment policies defined in \Cref{subsubsec:case_study_baselines}.}}
\label{fig:comparison_results_lowest}
\end{figure*}

\Cref{fig:comparison_results_lowest} shows the results for the lowest-demand weeks. In this regime, more robots are idle for longer periods, so even simple policies can often complete requests without rejection. Nevertheless, the proposed adaptive rollout policy still achieves the lowest median wait time for both the mean and $95$-th percentile metrics. The median improvement is approximately $2\%$ for daily mean wait time and approximately $7\%$ for daily $95$-th percentile wait time. At the $75$-th percentile, the proposed method reduces both metrics by approximately $10\%$ relative to the baselines.

The low-demand results show that the proposed method does not rely only on congestion to provide benefit. When capacity is abundant, the main opportunity is to position or preserve robots so that requests can be served quickly after they arrive. Prediction-aware methods can help in this setting, but only if the planner avoids overcommitting to uncertain predictions. The proposed rollout policy improves over both reactive baselines and the Idle Prediction baseline because it evaluates predicted demand through sampled future costs while keeping immediate assignments restricted to requests that have actually entered the system.

Across all demand regimes, the largest and most consistent gains appear in the $95$-th percentile wait-time metric. This suggests that the proposed adaptive rollout policy is particularly useful for reducing tail delays among serviced requests. Since most policies achieve similar service levels under the selected fleet size, these wait-time results provide the strongest evidence that prediction-aware rollout improves schedule quality beyond simply completing requests.

\subsubsection{Policy Planning-Time Results}
\label{subsubsec:case_study_runtime_results}

Finally, we evaluate the computational cost of each policy. This comparison is important because the proposed adaptive rollout policy performs substantially more computation than the reactive baselines: it generates robot-local candidate actions, evaluates those candidates over sampled future request scenarios, applies the interaction-aware base policy inside each rollout simulation, and scores the resulting simulated states. The candidate limit $L$, sample count $S$, one-robot-at-a-time decision rule, and decision-suppression mechanism described in \Cref{subsec:case_study_implementation_details} are therefore used to keep the computation compatible with online execution.

\begin{figure}
\centering
\includegraphics[width=\linewidth]{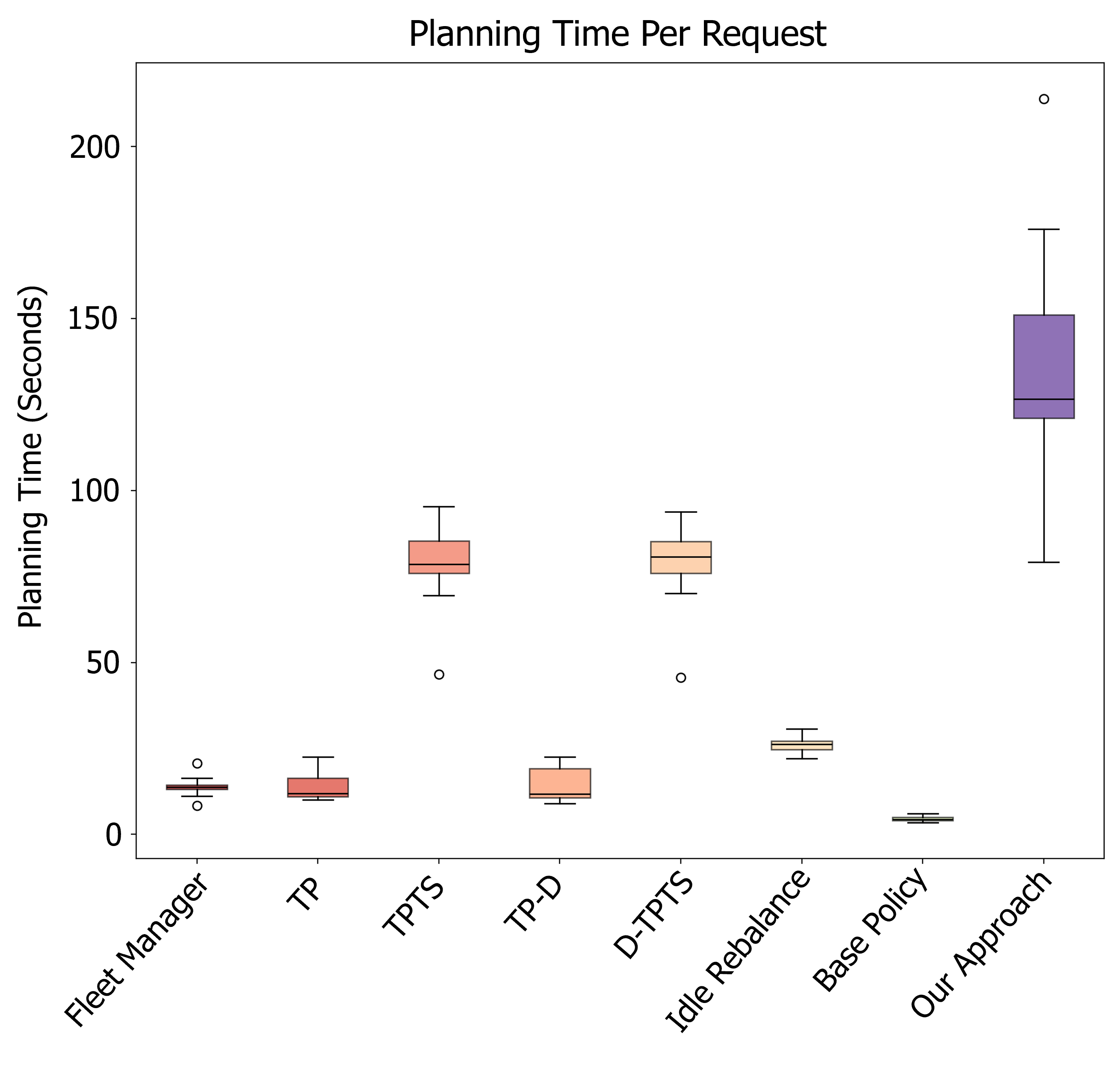}
\caption{\small{Planning time per request-triggered decision across the evaluated assignment policies. Each box plot summarizes the computation time required to process a decision during the evaluation runs. Lower values indicate faster online decision making. The proposed adaptive rollout policy requires more computation than myopic and token-passing baselines because it evaluates candidate assignments through sampled future scenarios, but its median planning time remains within the operational lead time of the case-study setting. Abbreviated policy labels correspond to the assignment policies defined in \Cref{subsubsec:case_study_baselines}.}}
\label{fig:planning_time_results}
\end{figure}

\Cref{fig:planning_time_results} reports the distribution of planning time required to process a request-triggered decision across the evaluated policies. As expected, the reactive and token-passing baselines require less computation because they make local assignment decisions without simulating sampled future demand. The proposed adaptive rollout policy has the largest planning time among the evaluated methods, but its median planning time remains approximately $2$ minutes per request-triggered decision. In the hospital case study, requests have a $5$-minute lead time between entry into the system and the earliest time at which they may be serviced. Thus, the observed median planning time remains within the operational window available for online decision making.

These results show that the improved wait-time performance of the proposed policy comes with a measurable computational cost. However, the runtime remains practical for the simulated hospital setting under the chosen rollout configuration. The implementation used for these experiments evaluates rollout simulations sequentially, even though many parts of the computation are naturally parallelizable. In particular, different sampled future scenarios, and in some cases different candidate-action evaluations, can be evaluated independently. A parallel implementation would therefore be expected to reduce wall-clock planning time without changing the policy logic. The reported runtimes should consequently be interpreted as a conservative estimate of the computational cost of the proposed rollout policy.

\subsection{Adaptive Rollout Ablation}
\label{subsec:case_study_ablation_studies}

The ablation study isolates the two adaptive mechanisms introduced in \Cref{subsec:adaptive_prediction_confidence,subsec:reoptimization}: adaptive prediction reweighting and selective re-optimization of unstarted assignments. All variants use the same one-robot-at-a-time rollout framework, the same candidate-generation procedure, the same future scheduled requests, and the same sampled prediction scenarios. The only difference across variants is whether predicted request costs are reweighted online and whether assigned but unstarted requests can be released for reconsideration.

We evaluate four rollout variants:
\paragraph{Rollout without adaptation.}
This variant uses future scheduled requests and sampled predicted requests in the rollout future-cost estimate, but prediction costs are not downweighted online and assigned but unstarted requests are not reconsidered.
\paragraph{Rollout with re-optimization only.}
This variant allows assigned but unstarted requests to be returned to the pending set when newly observed requests make the previous assignment order undesirable, but predicted request costs are not adaptively reweighted.
\paragraph{Rollout with prediction reweighting only.}
This variant updates the contribution of predicted requests to the rollout objective using recent forecast errors, but it does not release previously assigned unstarted requests.
\paragraph{Proposed adaptive rollout.}
This is the full proposed policy. It combines adaptive prediction reweighting with selective re-optimization of unstarted assignments.

\Cref{tab:ablation_results} reports the mean and standard deviation of request wait times for each demand regime. The mean captures average service responsiveness, while the standard deviation captures the spread of request delays. Lower values are better for both metrics.

\begin{table*}
\centering
\caption{Ablation study of the adaptive rollout components. All variants use the same rollout framework and include future scheduled requests in the future-cost estimate. The table reports the mean and standard deviation of request wait times, in seconds, for each demand level.}
\label{tab:ablation_results}
\small
\begin{tabular}{|p{0.32\textwidth}|cc|cc|cc|}
\hline
\multirow{2}{*}{Policy}
& \multicolumn{2}{c|}{High demand}
& \multicolumn{2}{c|}{Medium demand}
& \multicolumn{2}{c|}{Low demand} \\
\cline{2-7}
& Mean & Std.
& Mean & Std.
& Mean & Std. \\
\hline
Rollout without adaptation
& 49.25 & 90.02
& 34.23 & 78.21
& 25.65 & 68.72 \\ \hline

Rollout with re-optimization only
& 46.78 & 86.88
& 32.18 & 74.08
& 24.36 & 65.97 \\ \hline

Rollout with prediction reweighting only
& 49.39 & 90.49
& 33.70 & 77.25
& 25.17 & 67.69 \\ \hline

Proposed adaptive rollout
& \textbf{46.01} & \textbf{85.41}
& \textbf{31.88} & \textbf{73.54}
& \textbf{24.28} & \textbf{65.91} \\ \hline
\end{tabular}
\end{table*}

The full adaptive rollout policy achieves the lowest mean wait time and the lowest wait-time standard deviation in every demand regime. Relative to rollout without adaptation, the full policy reduces mean wait time by approximately $6\%$ under high demand, $7\%$ under medium demand, and $5\%$ under low demand. It also reduces the standard deviation of wait times by approximately $5\%$, $6\%$, and $4\%$, respectively. These reductions show that the adaptive mechanisms improve both average responsiveness and delay variability.

The ablation also shows that the two mechanisms play different roles. Re-optimization provides the larger individual improvement: by releasing assigned but unstarted requests, it allows the planner to repair schedules that become undesirable after new information arrives. Prediction reweighting alone provides smaller benefits and, under high demand, performs similarly to the non-adaptive rollout variant. This is expected because reweighting changes the influence of future predicted requests prospectively, but it cannot by itself revise assignments that have already been made. When reweighting and re-optimization are combined, the policy obtains the best performance in all regimes, indicating that the two mechanisms are complementary: reweighting reduces the future influence of unreliable predictions, while re-optimization repairs eligible commitments made under earlier forecasts.

\begin{figure}
    \centering
    \includegraphics[width=\linewidth]{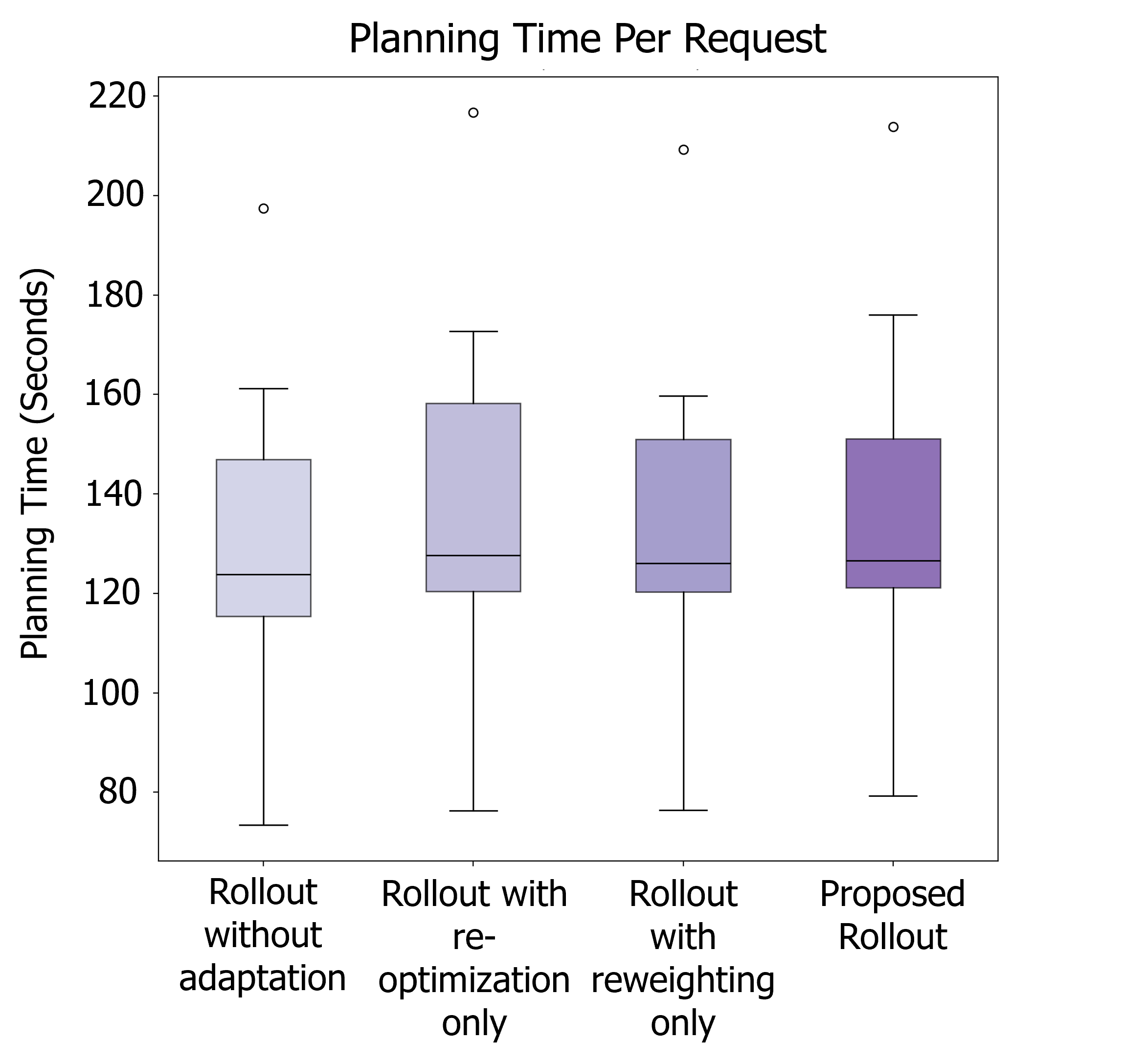}
    \caption{\small{Planning time per request-triggered decision across the ablation policies. Each box plot shows the distribution of computation time required to process a decision during the evaluation runs. Lower values indicate faster online decision making.}}
    \label{fig:ablation_runtimes}
\end{figure}

\Cref{fig:ablation_runtimes} compares the planning-time distributions for the rollout ablation variants. The adaptive mechanisms add only modest overhead relative to rollout without adaptation, because the dominant computational cost is evaluating candidate actions over sampled future scenarios. Prediction reweighting requires only lightweight updates to prediction-confidence weights, and re-optimization changes the set of eligible pending requests only when the trigger condition is satisfied.

The runtime results also show that the full adaptive rollout policy does not incur the largest tail planning times among the ablation variants. In particular, combining reweighting with re-optimization can reduce unnecessary schedule repair by lowering the influence of unreliable predictions before they lead to poor assignments. Thus, the full method improves wait-time performance without substantially increasing the computational cost of the underlying rollout procedure.

\section{Conclusion}
\label{sec:conclusion}

This paper studied online heterogeneous multi-robot task assignment and scheduling with scheduled requests, real-time requests, service-window constraints, and uncertain future demand. We formulated the problem as a finite-horizon stochastic dynamic program in which controls update the assignments and schedules of known requests, while future real-time requests enter through a stochastic disturbance process. The formulation explicitly distinguishes between the true deployment distribution $\mathbb P$ and the learned predictive distribution $\widehat{\mathbb P}$, which makes it possible to reason about prediction error and distribution shift within the online decision-making problem.

To address this problem, we proposed a prediction-aware adaptive rollout framework. The key design choice is to separate immediate commitments from future-demand estimates: robots can only be assigned to requests that have actually entered the system, while predicted requests influence decisions through sampled rollout simulations. This allows the planner to account for the opportunity cost of current assignments without prematurely committing robots to requests that may never occur. The rollout policy is made computationally tractable using one-robot-at-a-time decision making, heuristic candidate pruning, wait actions, and an interaction-aware base policy for future-cost estimation.

The framework also includes adaptive mechanisms for robustness under prediction error. Prediction-confidence weights are updated online from recent forecast errors and used to reduce the contribution of unreliable predicted requests in the rollout objective. Selective re-optimization complements this prospective correction by allowing assigned but unstarted requests to be reconsidered when new observations make the current schedule undesirable. Together, these mechanisms allow predictions to guide online allocation when they are useful, while limiting their influence when they become misleading.

The hospital-floor case study demonstrated how the framework can be instantiated in a realistic heterogeneous service setting. Historical request data were used both to train temporal point process models for future-request scenario generation and to select a fixed heterogeneous team composition before deployment. The empirical results showed that the selected team composition was sufficient to avoid persistent overload on the held-out floor-days, and that the proposed adaptive rollout policy reduced request wait times relative to reactive, token-passing, prediction-positioning, and myopic greedy baselines. The largest improvements appeared in tail wait-time metrics, indicating that prediction-aware rollout is especially useful for avoiding severe delays. The ablation study further showed that re-optimization and prediction reweighting play complementary roles: re-optimization repairs eligible commitments after new information arrives, while reweighting reduces the future influence of unreliable predictions.

Several directions remain for future work. First, the current implementation uses a finite set of sampled future scenarios and a heuristic base policy for downstream cost estimation. Future work could study stronger scenario-reduction methods, learned value approximations, or uncertainty-aware base policies that preserve rollout quality while reducing runtime. Second, the prediction-confidence mechanism currently relies on count-based forecast errors within predefined contexts; richer calibration methods could account for spatial structure, task dependencies, and correlations between local request processes. Third, the selective re-optimization rule could be extended to reason about the cost of schedule disruption more explicitly, especially in settings where changing unstarted assignments affects human workflows or resource availability. Finally, future deployments should evaluate the framework in closed-loop physical or high-fidelity hospital operations, where communication delays, navigation uncertainty, human interaction, and operational constraints may further affect the value of prediction-aware decision making.

Overall, the proposed framework provides a practical way to use learned future-demand predictions in online heterogeneous multi-robot scheduling. By restricting immediate actions to observed requests while adaptively weighting predicted requests in the lookahead objective, the method exploits useful forecasts without allowing unreliable predictions to dominate online decisions.

\section{Acknowledgments}
Authors are grateful for the Amazon Gift that funded this research in part. This work was also supported in part by the Defense Advanced Research Projects Agency (DARPA) under Grant No. $D24AP00319-01$. The views and conclusions expressed in this paper are those of the authors and do not reflect the official policy or position of the U.S. Army, U.S. Department of War, or U.S. Government. 

\printcredits

\bibliographystyle{cas-model2-names}

\bibliography{cas-refs}

\newpage

\bio{headshots/Garces_Headshot_4x5}
Daniel Garces is a Computer Science Ph.D. student in the School of Engineering and Applied Sciences at Harvard University, advised by Prof. Stephanie Gil. His research focuses on the development of model-based multi-agent reinforcement learning algorithms for task allocation in real world applications. He is interested in adaptation mechanisms and task allocation problems under uncertainty. He received his Bachelor’s degree in Computer Engineering from Columbia University in 2021.
\endbio

\bio{headshots/Castro_Headshot_4x5}
Sara Castro, MD, MPH, is an OB/GYN resident in the Harvard Mass General Brigham program. Her work focuses on patient perspectives and preferences regarding the use of robotics and emerging technologies in healthcare, with particular interest in equity, communication, care navigation, and language access. She earned her MPH in Healthcare Management, with a concentration in Public Health Leadership, from the Harvard T.H. Chan School of Public Health and her MD from Harvard Medical School. She also holds a BA in Medicine, Literature, and Society from Columbia University.
\endbio

\bio{headshots/Haimovich_Headshot_4x5}
Dr. Adrian Haimovich is an Assistant Professor of Emergency Medicine at Harvard Medical School and Director of the Division of Artificial Intelligence in the Department of Emergency Medicine at Beth Israel Deaconess Medical Center in Boston, Massachusetts. His research is at the intersection between emergency medicine and healthcare AI.
\endbio

\bio{headshots/Crowe_Headshot_4x5}
Dr. Byron Crowe is Chief Medical Officer at Doctronic, where he leads clinical strategy and development of its AI-native care model. He is also a Clinical Assistant Professor of Medicine at Stanford, focused on improving complex systems at scale. Previously, he served as CMO at Solera Health and on the faculty of Harvard Medical School, where he studied AI in clinical reasoning and health policy. Named to Modern Healthcare’s 2024 “40 Under 40,” he earned his MD from Emory and a master’s in clinical informatics management from Stanford, and is board-certified in internal medicine and clinical informatics.
\endbio

\bio{headshots/Gil_Headshot}
Stephanie Gil is an Assistant Professor in the Computer Science Department at Harvard University's School of Engineering and Applied Sciences, where she directs the Robotics, Embedded Autonomy, and Communication Theory (REACT) Lab. Previously, she served as an Assistant Professor at Arizona State University. Her research explores multi-robot systems, focusing on how communication and information exchange impact resilience and trusted coordination. She has received several recognitions, including the NSF CAREER Award (2019), ONR Young Investigator Program Award, DARPA Young Faculty Award, and was named 2020 Alfred P. Sloan Fellow. Stephanie earned her PhD from the Massachusetts Institute of Technology in 2014.
\endbio

\end{document}